\documentclass[10pt,journal,compsoc]{IEEEtran}
\ifCLASSOPTIONcompsoc
  \usepackage[nocompress]{cite}
\else
  \usepackage{cite}
\fi
\ifCLASSINFOpdf
\else
\fi

\usepackage{algorithm}
\usepackage{algorithmicx}
\usepackage{algpseudocode}
\usepackage{graphicx}
\usepackage{url}
\usepackage{amsmath}
\usepackage{amsfonts}
\usepackage{xcolor}
\usepackage{subfig}
\usepackage{boldline}
\usepackage{tablefootnote}
\usepackage{multirow}

\newcommand{\putvideo}{supplemental video\footnote{\url{https://youtu.be/s8VARNpBpSg}}}
\newcommand{\putcode}{at \url{https://github.com/donamin/llc}}

\begin{document}
%
\title{Learning Task-Agnostic Action Spaces for Movement Optimization}
%
%
%
%

\author{Amin~Babadi,
        Michiel~van~de~Panne,
		C.~Karen~Liu,
        and~Perttu~H{\"a}m{\"a}l{\"a}inen
\IEEEcompsocitemizethanks{\IEEEcompsocthanksitem Babadi and Hämäläinen are with the Department of Computer Science at Aalto University, Finland.\protect\\
E-mail: amin.babadi@aalto.fi, perttu.hamalainen@aalto.fi
\IEEEcompsocthanksitem Michiel van de Panne is with the Department of Computer Science
at University of British Columbia, Canada.\protect\\
E-mail: van@cs.ubc.ca
\IEEEcompsocthanksitem C. Karen Liu is with the Department of Computer Science
at Stanford University, USA.\protect\\
E-mail: karenliu@cs.stanford.edu}
}

%
%

\markboth{IEEE Transactions on Visualization and Computer Graphics,~Vol.~?, No.~?, August~2020}%
{Babadi \MakeLowercase{\textit{et al.}}: Learning Task-Agnostic Action Spaces for Movement Optimization}
%



\IEEEtitleabstractindextext{%
\begin{abstract}
We propose a novel method for exploring the dynamics of physically based animated characters, and learning a task-agnostic action space that makes movement optimization easier. Like several previous papers, we parameterize actions as target states, and learn a short-horizon goal-conditioned low-level control policy that drives the agent's state towards the targets. Our novel contribution is that with our exploration data, we are able to learn the low-level policy in a generic manner and without any reference movement data. Trained once for each agent or simulation environment, the policy improves the efficiency of optimizing both trajectories and high-level policies across multiple tasks and optimization algorithms. We also contribute novel visualizations that show how using target states as actions makes optimized trajectories more robust to disturbances; this manifests as wider optima that are easy to find. Due to its simplicity and generality, our proposed approach should provide a building block that can improve a large variety of movement optimization methods and applications.
\end{abstract}

}

\maketitle

\section{Introduction}
Movement optimization of physically simulated characters is common in robotics and computer animation. Although previous work has shown that a wide variety of movements can be generated through optimization, both in trajectory and policy optimization settings \cite{tassa2012synthesis,naderi2017discovering,bergamin2019drecon,2018-TOG-deepMimic,won2020scalable}, optimization is still prohibitively slow for many applications. This is largely due to the high dimensionality of the state and action spaces, as well as the complexity of movement dynamics plagued with contact discontinuities, which prohibits closed-form solutions. 

One way to make movement optimization easier is by using novel action parameterizations. Previous work has shown that the choice of action space can have a significant impact on the performance of movement optimization \cite{peng2017learning}. Learned or designed action spaces can also capture the task-invariant behavioral representations that can be re-used across different tasks, speeding up the problem solving process \cite{Merel2020Deep}. This idea has been long studied in the context of Hierarchical Reinforcement Learning (HRL), which has shown promising results in physically based character control \cite{peng2017deeploco,levy2019learning,lee2019scalable}. In HRL, the actions output by a high-level controller (HLC) are converted into low-level simulation or robot actuation commands using a low-level controller (LLC). For example, the muscle-actuated human simulation system of Lee et al. \cite{lee2019scalable} optimizes controls in the space of joint target accelerations, which are then converted to muscle actuations by a separately trained neural network. However, learning such control hierarchies can be computationally expensive and they are typically problem-dependent. A question remains whether there are \textit{truly generic action parameterizations or LLCs that would only need to be trained once}, but would still make movement optimization and learning easier across a wide range of tasks. 

In this paper, we investigate a promising candidate for a generic task-agnostic action space: \textit{We define HLC actions as target states to be reached by an LLC}. This poses no limitations on the optimized movements, as any movement can be described as a sequence of state variables such as body poses and root translations and rotations. This can also be motivated through human visuomotor control and movement pedagogy, as complex movement skills such as gymnastics are typically taught through demonstrations---visualizations of target state sequences---instead of explaining movements through low-level actions such as which muscles to contract and when. Hence, reaching and maintaining desired state variable values can be considered a central human meta-learning or ``learning to learn'' skill. Furthermore, recent research indicates that parameterizing actions as target states can improve both convexity and conditioning of movement optimization \cite{hamalainen2020visualizing}. However, the analysis of \cite{hamalainen2020visualizing} was limited to an inverted pendulum, in which case a simple P-controller suffices as the LLC. Generalization of the results to neural network LLCs and more complex agents was not demonstrated, calling for further research.   

\begin{figure*}[ht]
\centering
{\includegraphics[width=1\linewidth]{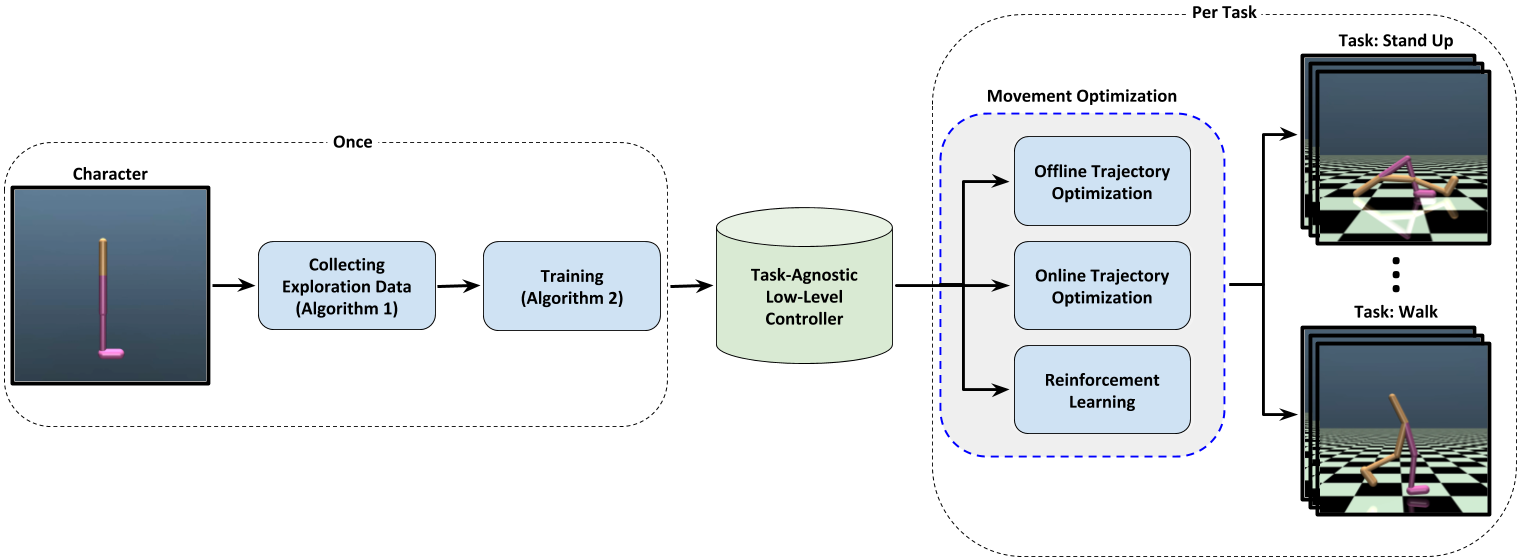} }
\caption{The system pipeline consists of three main steps: 1) collecting random exploration data to discover feasible states and actions of the simulation environment, 2) using the exploration data to train low-level controller, and 3) movement optimization using either offline/online trajectory optimization or reinforcement learning. The first two steps are task-agnostic and only need to be completed once for each environment. } 
\label{fig:system_overview}
\end{figure*}

This paper makes the following contributions:

\begin{itemize}
\item We propose a novel random exploration scheme for generating highly diverse training data for task-agnostic state-reaching LLCs, without dependencies on pre-recorded reference data. This makes our approach suitable for a wide variety of simulated characters and movements.  
\item Through extensive simulation experiments, we show that our LLCs improve movement optimization across multiple complex agents and multiple optimization methods including offline trajectory optimization, online trajectory optimization, and reinforcement learning (policy optimization). Our data also indicates that LLCs trained with the proposed exploration approach perform better than LLCs trained with simple random exploration data.
\item We visualize the resulting optimization landscapes with and without the LLC, extending the investigation of \cite{hamalainen2020visualizing} to complex agents and learned LLCs. Our results provide support for the improved convexity and conditioning suggested by \cite{hamalainen2020visualizing}. 
\end{itemize}

An overview of our system and approach is shown in Fig. \ref{fig:system_overview}. To augment the quantitative data presented in Section \ref{sec:experiments}, examples of the exploration behaviors and movement optimization results are included in the \putvideo. To the best of our knowledge, no previous work has conducted such comprehensive experiments on a task-agnostic action space that helps both trajectory and policy optimization. Our results indicate that an LLC like ours should provide a useful basis for building any system that utilizes movement optimization. All of the implementations used in this work can be found \putcode.


\section{Related Work}
Below, we review relevant previous work in the related areas of action space engineering, trajectory optimization, reinforcement learning, and movement exploration. 

\subsection{Action Space Engineering}\label{subsec:action_space_engineering}
One of the earliest successful examples of efficient action spaces is the Simple Biped Locomotion Control (SIMBICON) \cite{yin2007simbicon}. The control strategy of SIMBICON includes a Finite State Machine (FSM) whose states correspond to different phases of a biped walking cycle. This controller is able to produce robust locomotion movements using a small number of parameters, which can be tuned either manually or using motion capture data. This action parameterization has shown to be expressive enough to synthesize novel movements through interpolation and extrapolation \cite{Yin08}. SIMBICON's control strategy has also been extended to muscle-based control settings to synthesize high-quality locomotion gaits for bipedal creatures \cite{Geijtenbeek2013}. The latter uses Covariance Matrix Adaptation Evolution Strategy (CMA-ES) \cite{hansen2006cma} to optimize the controller parameters.

\subsection{Trajectory Optimization}
Parameterizing action sequences as splines is a modified action space common in synthesizing physically-based movements through trajectory optimization, i.e., optimizing actions to maximize or minimize an objective function that encodes movement goals as a function of both the action sequence and the resulting state-space movement trajectory. A spline parameterization reduces the problem dimensionality by expressing a long action sequence using only a few control points and interpolating between them. Plus, it enforces coordination across body joints (similar to SIMBICON \cite{yin2007simbicon}), which leads to more natural and smooth movements.

In offline settings, spline parameterization has been used in Contact-Invariant Optimization (CIO), a method that simultaneously optimizes the contact and behavior in different phases of the movement \cite{mordatch2012discovery}. In another work, it has been used for building a low-level controller that uses CMA-ES for synthesizing humanoid wall climbing movements \cite{naderi2018learning}.

In online settings, splines and sequential Monte Carlo sampling have been used for synthesizing interactive humanoid movements \cite{hamalainen2014online}. In a similar work, CMA-ES empowered by two seeding techniques was used to generate interactive martial arts movements for upper-body humanoid characters \cite{babadi2018intelligent}.

\subsection{Reinforcement Learning}
Finding novel action spaces is more widely studied in reinforcement learning than in trajectory optimization. This dates back at least to the early 1990s with Feudal Reinforcement Learning, that proposed solving the RL problem in multiple resolutions simultaneously \cite{dayan1993feudal}. The idea was then extended to temporally extended sequence of actions, also known as the Options framework \cite{sutton1999between}. The more general form is now known as Hierarchical Reinforcement Learning (HRL) \cite{dietterich2000hierarchical}. In HRL, a Low-Level Controller (LLC) is used for handling atomic actions, and policy optimization is applied to produce a High-Level Controller (HLC) that controls the agent via interaction with the LLC. The LLC can be either designed or built using reinforcement learning or other optimization methods.

Inspired by SIMBICON \cite{yin2007simbicon}, phase-based FSMs are one of the most popular action parameterizations in reinforcement learning. They have been used to learn dynamic terrain traversal policies for various 2D characters \cite{2015-TOG-terrainRL}. The same approach has been extended to Mixture of Actor-Critic Experts (MACE) to increase the training speed and enable expert specialization \cite{2016-TOG-deepRL}.

Another HRL approach is to define HLC as a gating network whose job is to choose among a number of low-level primitive controllers. In this setting, each primitive controller is specialized in performing a specific behavior (e.g., walking forward) or controlling a specific subset of joints (e.g., upper body). This approach has been used in \cite{liu2017learning}, where Deep Q-Learning \cite{mnih2015human} is used to learn a scheduler (i.e., HLC) that chooses one of the many control fragments (i.e., LLCs) to control the character for the next 0.1 seconds. Another similar work, called Multiplicative 
Compositional Policies (MCP), first trains several LLCs to imitate several short motion capture clips, and then trains a HLC to, at each timestep, compute a weighted composition of the LLCs' outputs \cite{MCPPeng19}.

In some HRL studies, the HLC is used to specify the long-term high-level goals that the LLC needs to achieve. A good example in this category is called DeepLoco, where the HLC is responsible for planning next footsteps and the LLC applies necessary low-level actions for satisfying the plan \cite{peng2017deeploco}. Another work has used end-to-end representation learning to find near-optimal goal spaces for continuous tasks \cite{nachum2018near}.

A popular HRL approach is to use a LLC that approximates inverse dynamics, i.e., outputs low-level actions required to take the agent to some desired state. Hindsight Experience Replay (HER) \cite{andrychowicz2017hindsight} and its recent hierarchical extension \cite{levy2019learning} use this idea by augmenting the experience replay dataset such that all reachable states can be considered as potential goal states. Another example is HIRO, a method that trains both HLC and LLC concurrently using off-policy experiences \cite{nachum2018data}. 

Several studies use reference animations to train robust control policies that demonstrate natural movement. One study uses adversarial imitation learning to train the LLCs, which are then used in RL settings to produce human-like motions \cite{merel2017learning}. A similar approach is used to learn control policies for quadruped characters \cite{luo2020carl}. In another study, LLC policies are trained to replicate short movement primitives of length $0.1$ to $0.3$ seconds, and then a HLC is trained to select among those LLCs based on visual input \cite{merel2018hierarchical}. In a similar study, a single LLC is built by training an autoencoder on multiple expert policies \cite{merel2020catch}. Another study trains general control policy by compressing a large number of expert policies that replicate motion capture moves \cite{merel2018neural}. A recent work uses a large corpora of motion capture clips to learn a low-level stochastic embedding space, which is then used in different tasks \cite{hasenclever2020comic}. A recent stream of research trains a kinematic motion generator on top of a state-reaching LLC to synthesize high-quality humanoid animation \cite{Park:2019,bergamin2019drecon, won2020scalable}.  However, these systems need motion capture data for training and the synthesized movements are goal-conditioned variations of the data. In contrast, our focus is on training the LLCs and optimizing movement without reference data, using a simple but efficient and task-agnostic exploration method. At least in principle, this should impose less limitations on the creativity and diversity of the results. 

It should be noted that reinforcement learning and trajectory optimization can also be combined. For example, \cite{levine2013guided,mordatch2014combining,rajamaki2017augmenting} use trajectory optimization to guide the training of a neural network policy. Trajectory optimization can also be used to generate states and actions for initializing an RL algorithm \cite{won2017train} or to synthesize reference movement trajectories for imitation learning using RL \cite{babadi2019self}.

\subsection{Exploration}
Using exploration to handle the uncertainty is one of the building blocks of optimal control \cite{thrun2002probabilistic}. In order to train a goal-conditioned low-level controller, one needs a proper dataset that includes the movements of interest and covers relevant regions of the state space. In the absence of a supervised dataset, the only way is to use some exploration approach to create one. Such random exploration is common in model-based reinforcement learning, where the exploration data is used to learn a forward dynamics model \cite{Kaiser2020Model,xie2016model,boney2019regularizing}.

Exploration is typically formulated as an intrinsically motivated optimization/learning problem, with a reward function that simulates the psychological mechanisms driving learning and exploration in biological agents, e.g., a drive to seek and experience novel or unpredictable stimuli. One implemention of this is to maximize the undiscounted information gain \cite{orseau2013universal}. Another example is called TEXPLORE-VANIR, a method whose reward encourages exploration when the model is uncertain or a novel experience is possible \cite{hester2017intrinsically}. A pseudo-count model based on density estimation has also been proposed to measure the novelty of encountered states in Atari 2600 environments \cite{bellemare2016unifying}. In a more recent work called Dreamer, exploration is done through imagination using a learned world model of latent state space \cite{hafner2019dream}. In this paper, we propose a simple random exploration approach that does not need novelty or density estimates, but still achieves much better coverage of possible agent states and produces more capable LLCs than naive exploration with random actions.


Recently, Sekar et al. \cite{sekar2020planning} have also conducted experiments in improving movement exploration, but using an approach largely orthogonal to ours. They proposed to train a world model using data produced through maximizing the disagreement among an ensemble of one-step forward models. Such forward models map an observation and action to the next observation, allowing model-based or ``imagination-based'' planning. In contrast, we learn inverse models that allow control optimization to operate in the space of the next observation(s). In addition, Sekar et al \cite{sekar2020planning} focused on reinforcement learning and relatively simple agents like 2D HalfCheetah and Pendulum, whereas we investigate both RL and trajectory optimization and include a 3D humanoid model in our experiments.

\section{Preliminaries}
Before going into the details of our approach, the following reviews the optimization methods and problem definitions we utilize.

\subsection{Trajectory Optimization}
Trajectory optimization denotes the process of searching for an optimal sequence of actions, which produce some desired movement trajectory when applied to a dynamical system. With differentiable dynamics, trajectory optimization can utilize gradient 
information \cite{tassa2012synthesis}, which is however unreliable with complex contact discontinuities. This can be overcome by sampling-based trajectory optimization \cite{hamalainen2014online,liu2012terrain} which works with any black-box dynamics simulator. The latter is the setting explored in this paper.

The core of sampling-based trajectory optimization consists of a simple iteration loop: 1) sample a number of random action trajectories and use them to simulate the state trajectories, 2) evaluate them using some cost or reward function, 3) use the best scoring trajectories to refine the sampling distribution for the next iteration. Finally, the best found trajectory is picked as the solution.

Trajectory optimization can be done offline or online. In the offline case, the starting state is fixed and the target is to find a single trajectory to achieve some goal. In online optimization case, also known as Model Predictive Control (MPC), when a trajectory is found, the simulation progresses by one timestep and the optimizer is asked to generate another trajectory starting from the next timestep. In other words, only the first action in the trajectory is actually executed and the rest is only used to compute the cost function and updating the sampling distribution. 

We simulate each sampled trajectory until a time horizon $T$ and the cost function is defined as the sum of costs over all timesteps. Suppose the character's state at the beginning of optimization is $\mathbf{s_\textit{t}} \in \mathcal{S}$ ($t=0$ in the case of offline optimization). Forward simulating a sequence of actions $\left\lbrace \mathbf{a_\textit{t}}, \mathbf{a_\textit{{t+1}}}, ..., \mathbf{a_\textit{{t+T-1}}} \right\rbrace \in \mathcal{A}^T$ results in a trajectory $\left( \mathbf{s_\textit{t}}, \mathbf{a_\textit{t}}, \mathbf{s_\textit{{t+1}}}, \mathbf{a_\textit{{t+1}}}, ..., \mathbf{s_\textit{{t+T-1}}}, \mathbf{a_\textit{{t+T-1}}}, \mathbf{s_\textit{{t+T}}} \right)$. Then, the problem will be to find the action sequence that minimizes the following accumulative cost:

$$
\sum\limits_{i=t}^{t+T-1} \left[ C_\mathcal{A}\left(\mathbf{a_\textit{i}}\right) + C_\mathcal{S}\left(\mathbf{s_\textit{{i+1}}}\right)\right]
,$$
where $C_\mathcal{A}$ and $C_\mathcal{S}$ are functions for computing the action and state costs, respectively. $C_\mathcal{S}$ usually encodes some information about the target movement. For example, if the goal is to produce walking movement, $C_\mathcal{S}$ can penalize the difference between the current and desired velocities, and $C_\mathcal{A}$ can penalize the amount of torques used in the simulation. This setup will encourage the character to move with the desired velocity while avoiding extreme movements \cite{Rajamaeki2018}.

\subsection{Reinforcement Learning}
In reinforcement learning, the learning process involves an agent interacting with an environment by observing a state, applying an action, and receiving a reward. The goal is to repeat this process and learn a \textit{policy}---a mapping of observed states to actions---that maximizes the accumulated reward over time \cite{sutton2018reinforcement}. Hence, the terms policy optimization and reinforcement learning are often used interchangeably.

In each timestep, the agent observes the current state $\mathbf{s_\textit{t}} \in \mathcal{S}$ and executes an action $\mathbf{a_\textit{t}} \in \mathcal{A}$ using a stochastic policy $\mathbf{a_\textit{t}} \sim \pi_\mathbf{\theta}\left(\mathbf{a_\textit{t}} \mid \mathbf{s_\textit{t}}\right)$, where $\mathbf{\theta}$ denotes policy parameters. In our case, the policy is a neural network and $\mathbf{\theta}$ denotes network weights. After that, the agent observes a scalar reward $r_\textit{t}$ along with the new state $\mathbf{s_\textit{{t+1}}}$. The goal is to find the optimal policy $\pi_{\mathbf{\theta^*}}\left(\mathbf{a_\textit{t}} \mid \mathbf{s_\textit{t}}\right)$ that maximizes the expected return, defined as follows:

$$
\mathbb{E}_{\mathbf{\tau} \sim \pi_\mathbf{\theta}\left(\mathbf{a} \mid \mathbf{s}\right)}\left[
\sum_{t=0}^{T} \gamma^{t}r_t
\right]
,$$
where $\mathbf{\tau}=\left( \mathbf{s_\textit{0}}, \mathbf{a_\textit{0}}, \mathbf{s_\textit{1}}, \mathbf{a_\textit{1}}, ..., \mathbf{s_\textit{{T-1}}}, \mathbf{a_\textit{{T-1}}}, \mathbf{s_\textit{{T}}} \right)$ is a trajectory generated by starting from $\mathbf{s_\textit{0}}$ (drawn from an initial state distribution) and following the policy $\pi_\mathbf{\theta}\left(\mathbf{a} \mid \mathbf{s}\right)$ afterwards. A discount factor $\gamma\in \left[0, 1\right]$ is used to ensure finite rewards as
$T\to\infty$.

In the case of continuous action spaces, policy gradient methods are a common optimization approach. At each optimization iteration, a number of episodes (simulated movement trajectories) are run up to the time horizon $T$ or until encountering a terminal state. The resulting states, actions, and rewards are then used to compute Monte Carlo gradient estimates, and gradient ascent is used to update the policy to improve the expected return.

In this paper, we evaluate our proposed action spaces using three popular policy gradient methods called \textit{Proximal Policy Optimization (PPO)} \cite{schulman2017proximal}, \textit{Soft Actor-Critic (SAC)} \cite{haarnoja2018soft}, and \textit{Twin-Delayed Deep Deterministic Policy Gradient (TD3)} \cite{fujimoto2018addressing}. PPO is an on-policy method---i.e., the episode actions are sampled from the policy being optimized---that uses the so-called clipped surrogate loss function or a KL-divergence penalty to allow large but stable policy updates per iteration. SAC is an off-policy method that tries to maximize the expected return while also maximizing policy entropy. TD3 is another off-policy algorithm that improves upon Deep Deterministic Policy Gradient (DDPG) \cite{lillicrap2015continuous} by using two value predictors, updating the policy less frequently than the value predictors, and adding noise to the target actions. PPO, SAC, and TD3 have shown to produce excellent results in computer animation and robotics \cite{peng2020learning,haarnoja2018soft,fujimoto2018addressing} and are now widely used in popular machine learning frameworks \cite{baselines,hafner2017tensorflow,juliani2018unity,pardo2020tonic}.

\textbf{\textit{Advantage estimation:}} A concept central to PPO is advantage estimation, which we modify in the PPO variant we use for LLC training (Section \ref{sec:method_training_llc}). The advantage of an action $\mathbf{a}$ in state $\mathbf{s}$ denotes how much the action improves the expected return, $A^\pi(\mathbf{a},\mathbf{s})=Q^\pi(\mathbf{a},\mathbf{s})-V^\pi(\mathbf{s})$, where $V^\pi(\mathbf{s})$ is the value function or expected return from state $\mathbf{s}$ following policy $\pi$, and $Q^\pi(\mathbf{a},\mathbf{s})=r(\mathbf{a},\mathbf{s})+\gamma V^\pi(\mathbf{s}')$ is the expected return of taking action $\mathbf{a}$ and then following the policy. The state $\mathbf{s}'$ is the next state resulting from taking action $\mathbf{a}$ in state $\mathbf{s}$. PPO  and other advantage-based policy gradient methods adjust the policy parameters to increase the probability of positive advantage actions, and decrease the probability of negative advantage actions. To compute the advantages, one usually trains a separate neural network to predict $V^\pi(\mathbf{s})$. This inevitably causes some bias, which is why PPO uses Generalized Advantage Estimation (GAE) \cite{schulman2015high}, a simple procedure that allows trading bias for variance.

\section{Simulation Environment} \label{sec:simulation_environment}

We base our experiments on four challenging MuJoCo \cite{todorov2012mujoco} environments (agents) in OpenAI Gym \cite{brockman2016openai}: \textit{HalfCheetah-v2}, \textit{Walked2d-v2}, \textit{Hopper-v2}, and \textit{Humanoid-v2}. The details of these environments are shown in Table \ref{TBL:characters_details}. In order to show how a trained LLC can work across tasks, we defined the following six tasks for each of the environments:

\begin{enumerate}
    \item \textit{Default}: The default locomotion task in OpenAI Gym MuJoCo environments, where agent receives rewards proportional to speed.
    \item \textit{Slow walk}: A locomotion task where the agent is rewarded for staying close to a standing pose while reaching a target velocity of $1m/s$ along the $x$ axis.
    \item \textit{Run}: Similar to \textit{slow walk}, but with a target velocity of $4m/s$.
    \item \textit{Back walk}: Similar to \textit{slow walk}, but using a target velocity of $-1m/s$.
    \item \textit{Balance}: Similar to \textit{slow walk}, but with the target speed of zero.
    \item \textit{Stand up}: The characters begin fallen on the ground, and the goal is to stand up.
\end{enumerate}

For tasks 1-5, we utilize the default OpenAI Gym termination criteria: Agents except HalfCheetah-v2 are considered to fail if they fall down, which causes an episode to terminate without reward. Task 6 episodes only terminate after a time limit of 10 seconds.

Task 1 uses the default OpenAI Gym reward function. For tasks 2-6, we define the reward function as:

$$
r = -\left\lVert \mathbf{s}_r - \mathbf{g}_r \right\rVert^2 - 0.01 \times \frac{\left\lVert \mathbf{a} \right\rVert^2}{N_{DOF}}
,$$
where $\mathbf{s}_r$ and $\mathbf{g}_r$ are subsets of current and target states used for the reward computation, and $\mathbf{a} \in \mathbb{R}^{N_{DOF}}$ denotes the low-level action (a vector of joint torques) applied in state $\mathbf{s}$. The full agent state $\mathbf{s}$ observed by RL algorithms comprises root vertical position, root velocity, root angular velocity, and joint angles and angular velocities. The subsets $\mathbf{s}_r,\mathbf{g}_r$ used for the reward computation comprise root velocity and joint angles. The target joint angles correspond to a default standing pose. This form of penalizing the deviation from a default pose is a common technique for preventing unnatural movements in optimization-based motion synthesis (e.g.,  \cite{Rajamaeki2018,babadi2018intelligent}). We believe that these tasks, despite their similarities, provide a wide range of challenges for movement optimization. Similar tasks have also been used in previous work on meta learning  \cite{heess2017emergence,gupta2018unsupervised}.

Although the simulation timesteps $\delta_t$ for common MuJoCo environments vary, we use a fixed action frequency of $f_a=10$Hz in all experiments, repeating each low-level action for $1/(f_a \delta_t)$ simulation steps. 

\begin{table}[!ht]
\caption{Details of the OpenAI Gym environments used in the experiments}
\begin{center}
\begin{tabular}{|c|c|c|c|c|}
\hline
\textbf{Environment}&\textbf{Bones}&\textbf{State variables}&\textbf{Action variables}\\
\hlineB{3.5}
\textbf{HalfCheetah-v2}&$7$&$17$&$6$\\
\hline
\textbf{Walker2d-v2}&$7$&$17$&$6$\\
\hline
\textbf{Hopper-v2}&$4$&$11$&$3$\\
\hline
\textbf{Humanoid-v2}&$13$&$45$&$17$\\
\hline
\end{tabular}
\label{TBL:characters_details}
\end{center}
\end{table}

\section{Low-Level Controller}
We train the LLCs in two steps. First, we use a novel contact-based random exploration method to discover the feasible states and actions of the simulation environment. Then, we utilize an iterative process of value estimation and LLC training. One of the key ideas behind our exploration scheme is resetting the simulation to diverse initial states. This utilizes the fact that in computer animation, as opposed to robotics, manipulating the states can be done without any extra cost. A good example of using this property, which also inspired our episode initialization, is Reference State Initialization of DeepMimic \cite{2018-TOG-deepMimic}. However, while DeepMimic uses randomly selected states from a motion capture database, our initialization is designed to cover all feasible movement states, without any need for motion data.

\subsection{Contact-Based Exploration} \label{subsec:contact-based-exploration}

\textbf{\textit{Problem Definition:}} We define the LLC training data generation as an exploration problem: collect data of states, actions, and resulting next states such that: 1) the data covers the joint space of states and actions as completely as possible, and 2) the data enables learning an LLC that allows the agent to efficiently actuate itself to transition between states. Since we are aiming to build task-agnostic LLCs, we refrain ourselves from using any task-dependent reward signals during exploration.

\textbf{\textit{Design Rationale:}} Our proposed solution focuses on \textit{discovering diverse contact configurations}. This is crucial because a character without an actuated root---e.g., a biped or a quadruped, as opposed to a robot arm mounted on a pedestal---can only affect its environment and actuate its center of mass through contacts. Although contacts are essential for balance and controlled motion, contact discontinuities are what makes learning, and modeling the dynamics hard \cite{mordatch2012discovery}. Thus, our working hypothesis is that an exploration method designed to discover diverse contact configurations should allow learning better LLCs. 

\begin{algorithm}[!t]
\caption{Contact-Based Random Exploration}\label{alg:contact_exploration}
\begin{algorithmic}[1]
\Function{Explore}{$K$}
\State \textbf{Input}: Rollout horizon $K$
\State \textbf{Output}: Exploration buffer $\mathcal{B}$
\State Initialize exploration buffer $\mathcal{B}\gets \left\lbrace \right\rbrace$
\While {iteration budget $N$ not exceeded}
\State $r \gets$ Random number in $[0, 1]$ \label{line:exp_init_y_begin}
\If{$r<p_{free}$}
\State $d_{ground} \gets$ Random number in $[0,h_{free}]$
\Else
\If{$r<p_{free}+p_{close}$}
\State $d_{ground} \gets$ Random number in $[0,h_{close}]$
\Else
\State $d_{ground} \gets 0$
\EndIf
\EndIf \label{line:exp_init_y_end}
\State Reset simulation to a random state and locate the character so that it is $d_{ground}$ units above the ground \label{line:exp_init}
\For {$t=0,1,...,K-1$} \label{line:exp_simulation_start}
\State Observe current state $\mathbf{s_\textit{t}}$
\State Sample random action $\mathbf{a_\textit{t}}$
\State $\mathbf{s_\textit{{t+1}}} \gets \Call{Simulate}{\mathbf{s_\textit{t}},\mathbf{a_\textit{t}}}$
\State $\mathcal{B} \gets \mathcal{B}\cup \left\lbrace[\mathbf{s_\textit{t}},\mathbf{a_\textit{t}},\mathbf{s_\textit{{t+1}}}]\right\rbrace$ \label{line:exp_simulation_end}
\EndFor
\EndWhile
\State \textbf{return} $\mathcal{B}$
\EndFunction
\end{algorithmic}
\end{algorithm}

\textbf{\textit{Exploration Algorithm:}} Our exploration method is detailed in Algorithm \ref{alg:contact_exploration}. The method uses simple random actions while still achieving diverse exploration through the following two principles:
\begin{enumerate}
\item The exploration episodes are short ($K=5$, corresponding to 0.5 seconds of simulation time) to reduce bias towards dynamics attractors such as falling down.
\item The initial state of each episode is randomized to maximize the diversity of contact configurations while ensuring physical plausibility.
\end{enumerate}


The initial state randomization proceeds as follows: First, a distance from ground $d_{ground}$ is randomly determined such that the character is either in the air at height $h_{free}$ with probability $p_{free}$, or close to the ground at height less than $h_{close}$ with probability $p_{close}$, or on the ground otherwise (Lines \ref{line:exp_init_y_begin}-\ref{line:exp_init_y_end}). We use $h_{free}=1$, $h_{close}=0.05$, $p_{free}=0.1$, and $p_{close}=0.4$ to bias the initialization towards states where the character is in contact with ground or likely to make contact with ground in the next few simulation steps. The rest of the initial state variables are chosen by uniformly selecting the rotation, velocity, and angular velocity, as well as joints angles and angular velocities (Line \ref{line:exp_init}), and adjusting character vertical position so that the distance to ground matches $d_{ground}$. The valid range for the joint angles and angular velocities is precomputed by keeping the character in the air and actuating the joints with random actions. As humans and many other moving agents spend most of their lives upright and in relatively slow movement, we bias the sampled root rotations and velocities by linearly interpolating towards a default upright pose by a uniformly sampled amount. Examples of the state initialization are shown in the supplementary video. 


\subsection{State Space Coverage}

In order to analyze how our contact-based exploration method in Algorithm \ref{alg:contact_exploration} covers the state space, we compared it against a naive random exploration baseline, where episodes are initialized using the default OpenAI Gym state initialization (an upright pose with zero intial velocity plus some random noise), and episodes are longer ($K=100$) to make it possible for the initially non-moving agent to gain velocity and reach diverse states. Such a random exploration approach is used in many recent papers \cite{chua2018deep,ghosh2019learning,schmeckpeper2019learning,boney2019regularizing}.

\begin{figure}[!t]
\centering
\subfloat[HalfCheetah-v2 with naive exploration]
{{\includegraphics[width=0.45\linewidth]{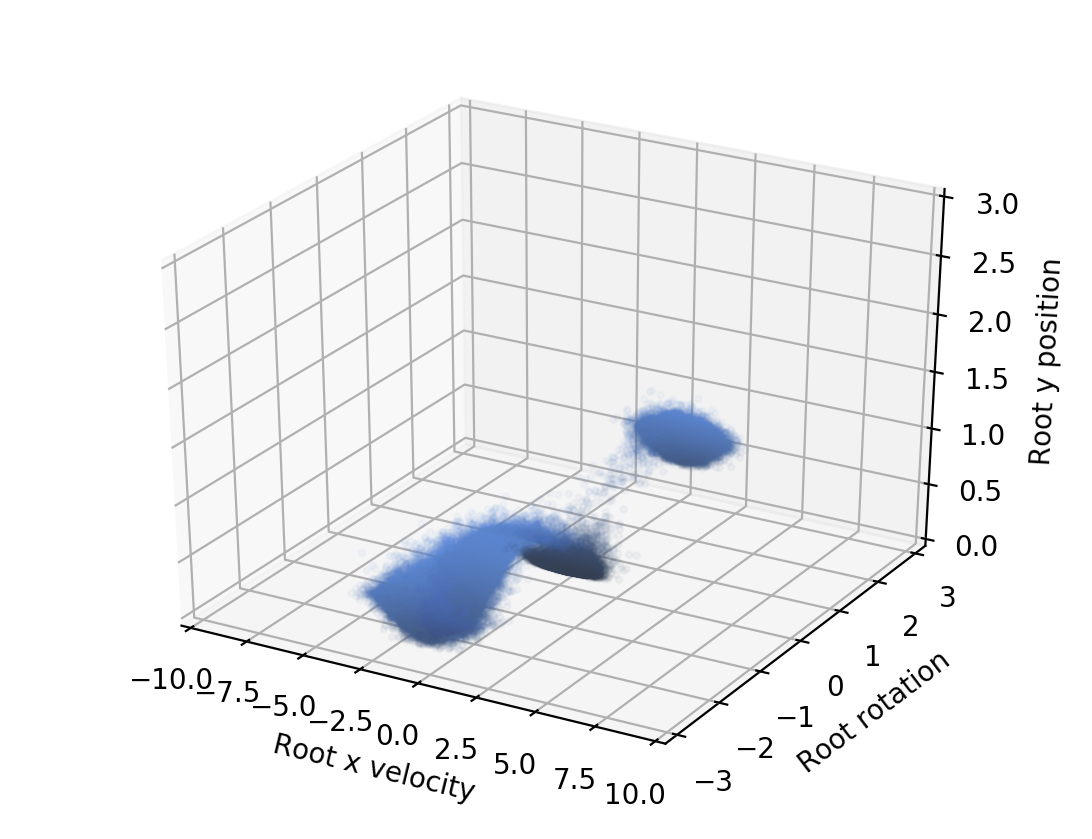} }\label{fig:exploration-halfcheetah-naive}}
\qquad
\subfloat[HalfCheetah-v2 with contact-based exploration]{{\includegraphics[width=0.45\linewidth]{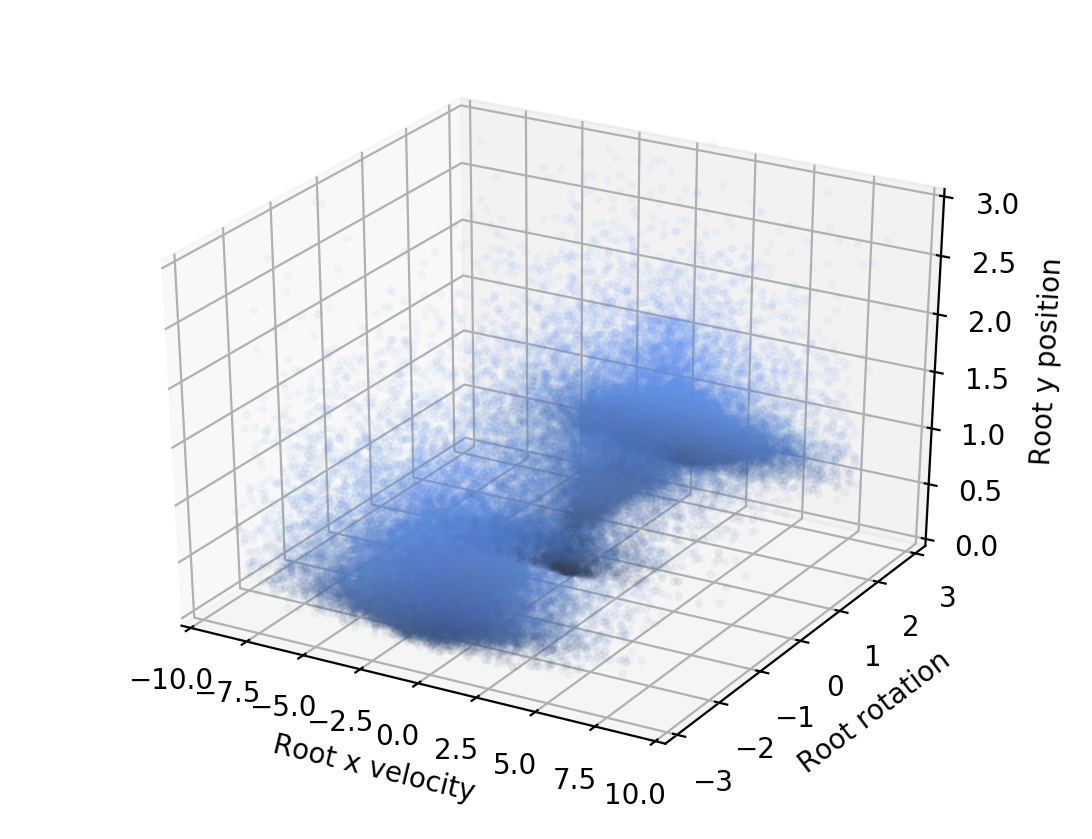} }\label{fig:exploration-halfcheetah-contact}}
\vfill
\subfloat[Hopper-v2 with naive exploration]
{{\includegraphics[width=0.45\linewidth]{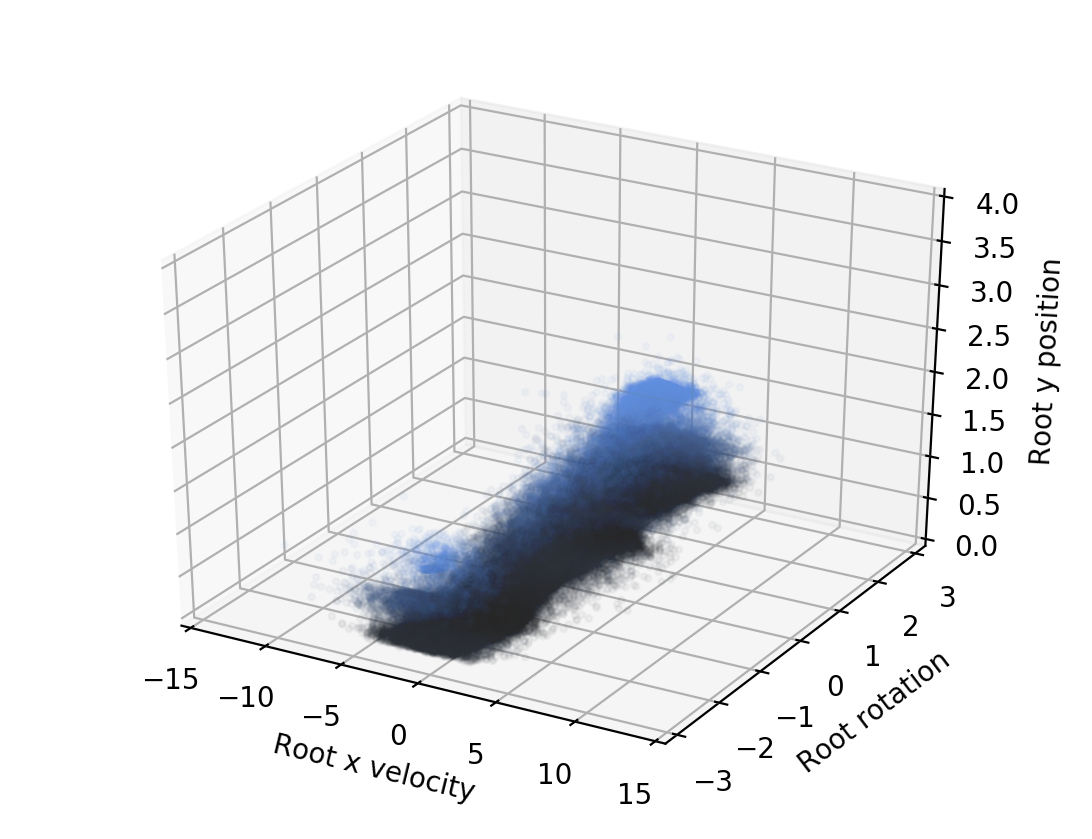} }\label{fig:exploration-hopper-naive}}
\qquad
\subfloat[Hopper-v2 with contact-based exploration]{{\includegraphics[width=0.45\linewidth]{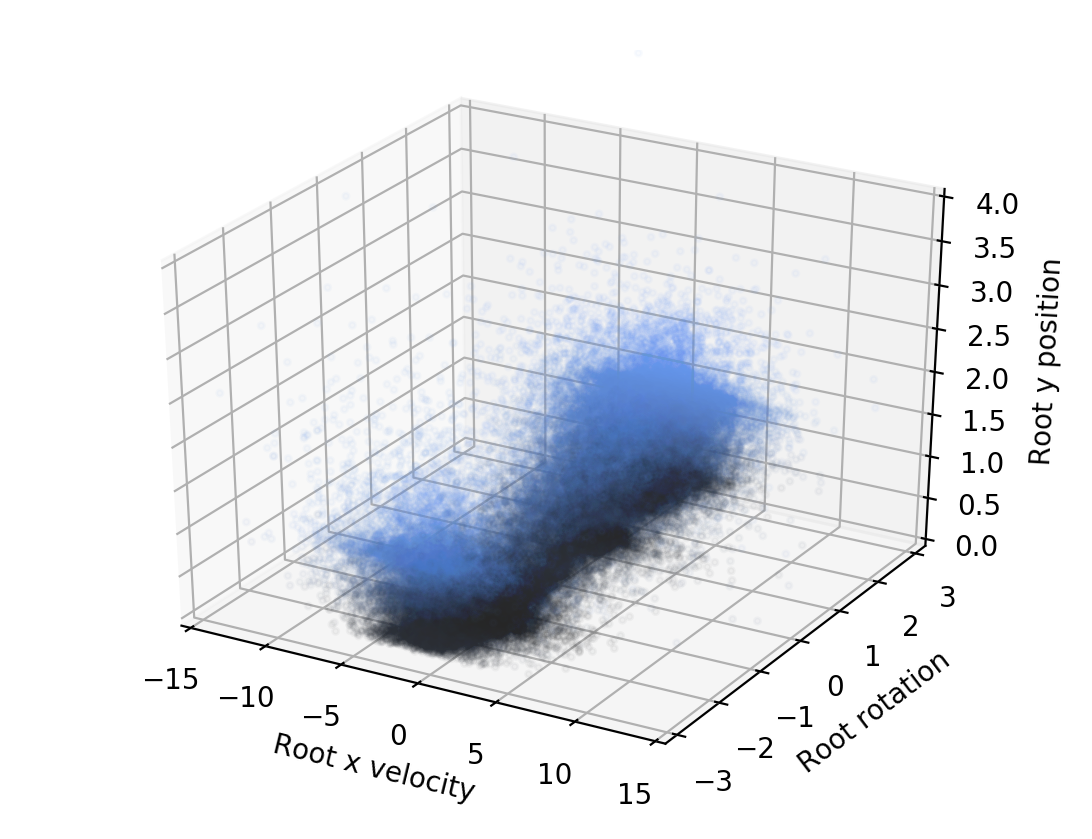} }\label{fig:exploration-hopper-contact}}
\vfill
\subfloat[Walker2d-v2 with naive exploration]
{{\includegraphics[width=0.45\linewidth]{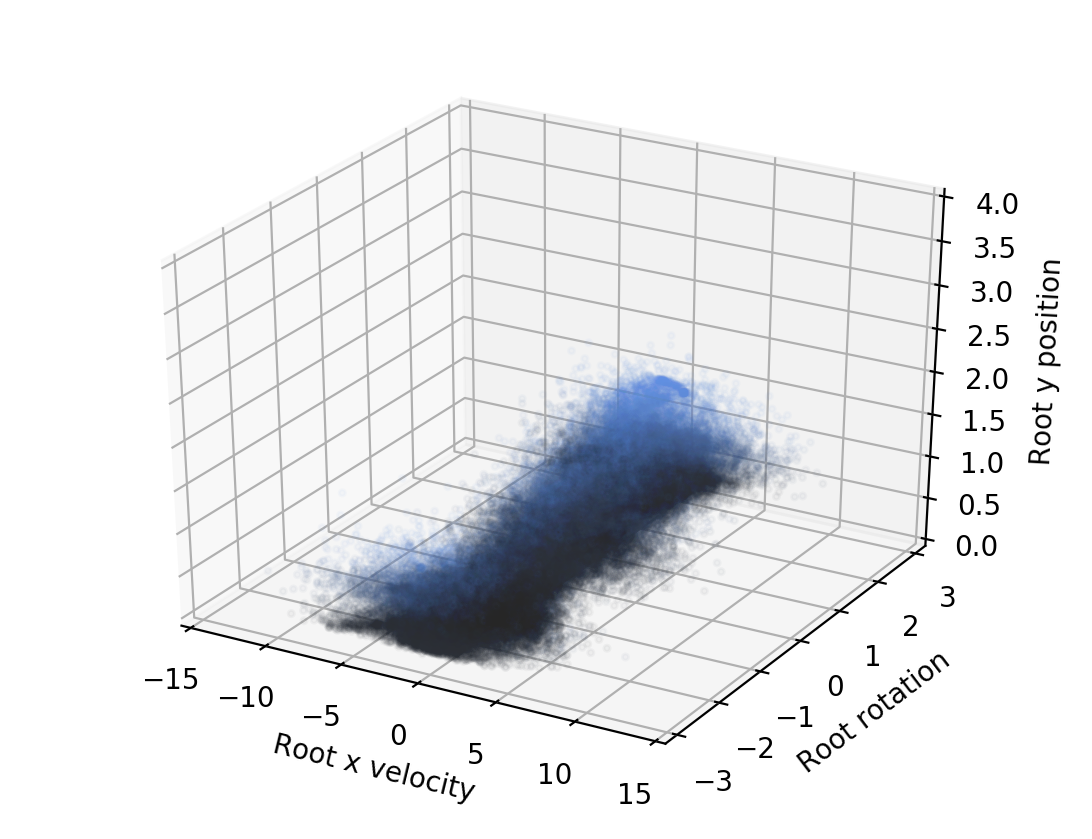} }\label{fig:exploration-walker2d-naive}}
\qquad
\subfloat[Walker2d-v2 with contact-based exploration]{{\includegraphics[width=0.45\linewidth]{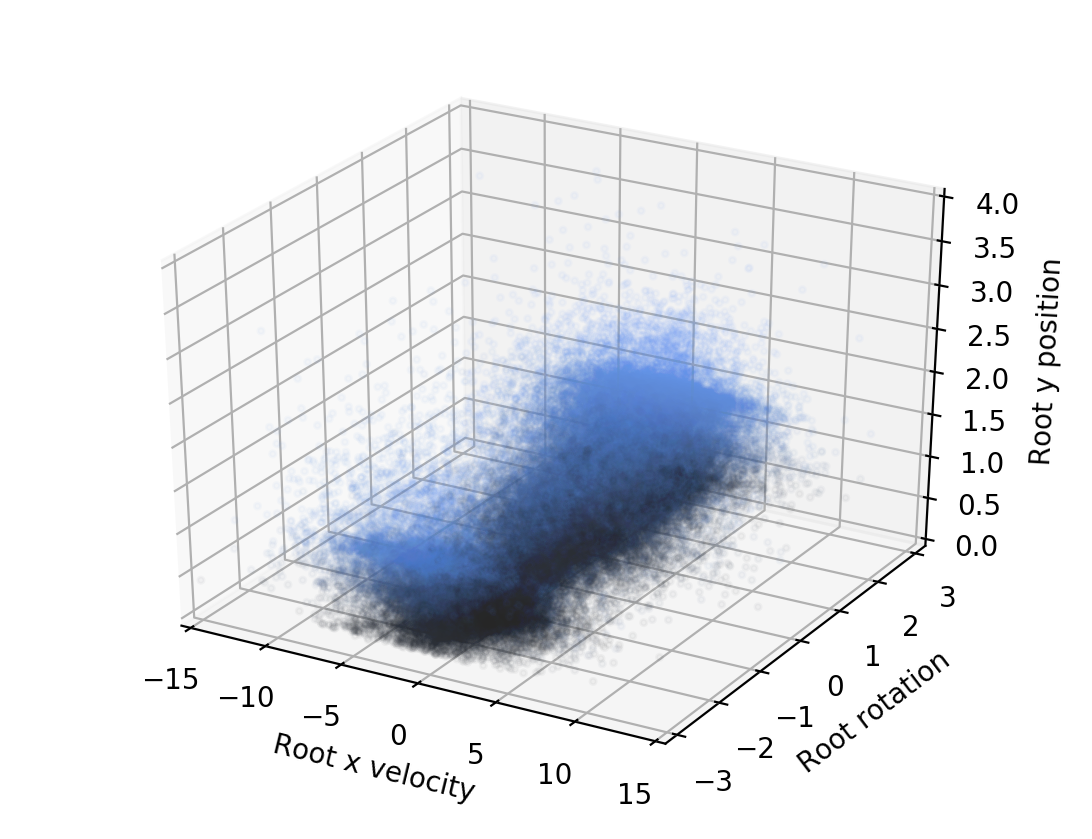} }\label{fig:exploration-walker2d-contact}}
\caption{Scatter plots of visited states when using naive exploration (left) and the proposed contact-based exploration (right). Standing and fallen states are shown in light and dark, respectively. Almost immediately after an episode starts, naive exploration causes the agents to fall down. Contact-based exploration however, uses short-length episodes with randomized state initialization to visit a more diverse set of states.}
\label{fig:exploration_scatter_plots}
\end{figure}

Figure \ref{fig:exploration_scatter_plots} shows scatter plots of the visited states using both methods in \textit{HalfCheetah-v2}, \textit{Walked2d-v2}, and \textit{Hopper-v2} environments. The plots visualize the $x$ velocity, rotation, and $y$ position of the root, observed in $100,000$ visited states. Upright and fallen states are shown in light and dark, respectively. Figure \ref{fig:exploration_scatter_plots} indicates that the naive random exploration achieves poor state space coverage, with the agent to wasting simulation budget in fallen states. In contrast, our contact-based exploration enables the agent to visit more diverse states.

\subsection{Training the Low-Level Controller}\label{sec:method_training_llc}

We use the exploration data to train a state-reaching LLC, denoted by the policy $\pi_\textit{H}\left(\mathbf{a}|\mathbf{s},\mathbf{G}\right)$ that allows sampling and action $\mathbf{a} \sim \pi_\textit{H}\left(\mathbf{a}|\mathbf{s},\mathbf{G}\right)$ for driving the agent to follow a trajectory of desired next states $\mathbf{G}$ of duration $H$ from the current state $\mathbf{s}$. Later in Section \ref{sec:experiments}, we also study the case where $\mathbf{G}$ only specifies a single target state that is $H$ steps into the future. However, our results show that the former is superior; thus throughout the text $\mathbf{G}$ is assumed to be a state trajectory, unless otherwise specified. Also note that in case of $H=1$, the LLC can be considered as an inverse dynamics controller. During LLC training, we use a task-agnostic reward that computes the negated squared deviation from the desired state(s).

We implement the LLCs using multi-layer perceptron (MLP) neural networks that take $\mathbf{s},\mathbf{G}$ as input, and output a mean and a diagonal covariance matrix for sampling the actions $\mathbf{a}$ from a Gaussian distribution conditioned on $\mathbf{s},\mathbf{G}$. We train a separate network for each $H=\left\lbrace 1,2,...,H_{max} \right\rbrace$, which we found to be more robust than using $H$ as an additional network input. We use $H_{max}=5$, corresponding to maximum target trajectory length of 0.5 seconds. 

\textbf{\textit{Overview:}} Our LLC training approach is a variant of the Proximal Policy Optimization (PPO) \cite{schulman2017proximal} designed according to the following principles:

\begin{itemize}
\item To prevent out-of-distribution problems and make LLC robust to anything an HLC might output (e.g., in initial HLC training when the target states output by the HLC may be random and infeasible), we initialize the LLC training episodes using our diverse exploration data, and use a mixture of both feasible and infeasible target state sequences. 
\item We train the LLC and run our experiments using multiple values for $H$, assuming that there exists a tradeoff between LLC simplicity and robustness: Low $H$ should allow easier learning and smaller LLC networks, but also reduces the set of states that the LLC can reach. We investigate the effect of $H$ in our experiments in Section \ref{sec:experiments}.
\item In training for multiple $H$ values, we assume that a target trajectory of $H$ states can be followed by taking a single action using $\pi_\textit{H}$, and then continuing with $\pi_\textit{H-1},\pi_\textit{H-2},...,\pi_\textit{1}$. As detailed below, this allows us to simplify PPO and estimate advantages without a value predictor network.
\end{itemize} 

The LLC training process is detailed in Algorithm \ref{alg:llc_training}. The main part is the \Call{TrainLLCs}{} function, which successively trains LLCs for $H=1,2,...,H_{max}$.  We train each new LLC in $M=500$ iterations, with the simulation budget of $N=15000$ actions per iteration ($7.5M$ actions in total). 

\textbf{\textit{Supervised pretraining:}} For each trained LLC $\pi_H$, we first pretrain in a supervised manner so that action mean and variance approximate the exploration data (Line \ref{line:llc_training_pretrain}). The pretraining uses all subsequences of $H+1$ states of each exploration episode, using the first subsequence state as the current state, first action as the ``ground truth'' correct action, and the rest of the states as the target sequence.

\textbf{\textit{Episodic training:}} After the pretraining, we continue with PPO, running on-policy episodes of $H$ actions. For each episode, we sample an initial state from the exploration data (Line \ref{line:llc_training_traj_init_start}), and target sequence of $H$ states (Line \ref{line:llc_training_traj_init_end}). The target state sequence sampling is designed to mostly provide feasible targets from the exploration data, but also include completely random targets to make the LLC robust for anything the HLC outputs and thus prevent out-of-distribution problems. More specifically, we use a mixture of three distributions:
\begin{itemize} 
\item With probability $p_{e}$, we use the state sequence following the initial state in the exploration data. We use $p_{e}=0.8$.
\item With probability $p_{l}$, we uniformly sample a final target state within the state variable ranges in the exploration data, and linearly interpolate the target sequence between the initial and the final state. We use $p_{l}=0.1$.
\item Otherwise, we use a constant target state sampled uniformly between the minimum and maximum of each state variable in the exploration data.
\end{itemize}
Appendix E provides an ablation study that compares the above to using only feasible target state sequences, i.e., $p_{e}=1.0,p_{l}=0$.

\begin{algorithm}[!t]
\caption{Training Low-Level Controllers (LLC)}\label{alg:llc_training}
\begin{algorithmic}[1]
\Function{TrainLLCs}{$H_{max}$}
\State \textbf{Input}: LLC horizon $H_{max}$
\State \textbf{Output}: The LLC policies $\pi_{1:H_{max}}$
\State $\mathcal{B}\gets$\Call{Explore}{$H_{max}$} \Comment{Algorithm \ref{alg:contact_exploration}}
\State Initialize $\pi_{1:H_{max}}$
\For {$H=1,2,...,H_{max}$}
\State Pretrain $\pi_\textit{H}\left(\mathbf{a}|\mathbf{s},\mathbf{G}\right)$ using exploration buffer $\mathcal{B}$ \label{line:llc_training_pretrain}
\For {iteration$=1,2,...,M$}
\While {iteration budget $N$ not exceeded} \label{line:training_llc_policy_evaluation_begin}
\State Sample state $\mathbf{s}$ from $\mathcal{B}$ \label{line:llc_training_traj_init_start}
\State Sample target state trajectory $\mathbf{G}_\textit{{1:H}}$ \label{line:llc_training_traj_init_end}
\For {value sample $i=1,2,...,N_{adv}$} \label{line:llc_training_inner_loop_begin}
\State $\mathbf{a_\textit{i}},Q_\textit{i} \gets \Call{CalcQ}{\mathbf{s},H,\mathbf{G}_\textit{{1:H}},\pi_{1:H_{max}}}$ \label{line:training_llc_recursive_value}
\EndFor \label{line:llc_training_inner_loop_end}
\State Estimate state value $V \gets \Call{Mean}{\left\lbrace Q_\textit{i}\right\rbrace}$ \label{line:llc_training_value_mean}
\For {value sample $i=1,2,...,N_{adv}$}
\State Estimate the advantages $A_\textit{i} \gets Q_\textit{i}-V$ \label{line:llc_training_advantage_estimation}
\EndFor
\EndWhile \label{line:training_llc_policy_evaluation_end}
\State Update $\pi_\textit{H}$ using advantages $\left\lbrace A_\textit{i}\right\rbrace$ and PPO \label{line:llc_training_ppo_update}
\EndFor
\EndFor
\State \textbf{return} $\pi_{1:H_{max}}$
\EndFunction

\State
\Function{CalcQ}{$\mathbf{s},H,\mathbf{G}_\textit{{1:H}},\pi_{1:H_{max}}$}
\State \textbf{Input}: Current state $\mathbf{s}$, LLC horizon $H$, target state trajectory $\mathbf{G}_\textit{{1:H}}$, LLC policies $\pi_{1:H_{max}}$.
\State \textbf{Output}: First executed action $\mathbf{a}$ when using $\pi_{1:H_{max}}$ to reach $\mathbf{G}_\textit{{1:H}}$ from $\mathbf{s}$, action value $Q\left(\mathbf{s},\mathbf{a}\right)$
\State Sample action $\mathbf{a} \sim \pi_\textit{H}\left(\mathbf{a}|\mathbf{s},\mathbf{G}_\textit{{1:H}}\right)$ \label{line:training_llc_simulate_begin}
\State $\mathbf{s'} \gets \Call{Simulate}{\mathbf{s},\mathbf{a}}$
\State Compute the reward $r \gets -\left\lVert \mathbf{s'} - \mathbf{G}_\textit{{1}} \right\rVert^2$ \label{line:llc_training_reward}
\If{$H>1$}
\State $a',Q' \gets  \Call{CalcQ}{s',H-1,\mathbf{G}_\textit{{2:H}},\pi_{1:H_{max}}}$
\State $Q \gets r + Q'$
\Else
\State $Q \gets r$
\EndIf \label{line:training_llc_simulate_end}
\State \textbf{return} $\mathbf{a},Q$
\EndFunction

\end{algorithmic}
\end{algorithm}

\textbf{\textit{Modified advantage estimation:}} As the target sequence length decreases over the episode steps, only the first episode action is actually sampled from the $\pi_H$ being trained, and the rest are determined by the previously trained shorter-horizon LLCs. As a consequence, only the first action of the episode is used for updating $\pi_H$, and the rest of the episode only affects the advantage estimate of the first action. For the first action, we can also directly compute the advantages as $A(\mathbf{a},\mathbf{s})=Q(\mathbf{a},\mathbf{s})-V(\mathbf{s})$: We run $N_{adv}=4$ episodes from each initial state (lines \ref{line:llc_training_inner_loop_begin}-\ref{line:llc_training_inner_loop_end}), using the episode return as a single-sample Monte Carlo estimate of $Q$, returned by the \Call{CalcQ}{} function, which recursively samples the actions and collects rewards using the previously trained LLCs until episode end (Line \ref{line:training_llc_recursive_value}). The rewards are calculated as the negated squared deviation from the target state (Line \ref{line:llc_training_reward}). The mean of the episode returns is used as the Monte Carlo estimate of $V(\mathbf{s})$ (Line \ref{line:llc_training_value_mean}). Thus, we do not need to train a value predictor network with the episode returns, simplifying the PPO algorithm.

When using a value predictor network, we experienced PPO diverging. A plausible explanation for this is that the highly diverse and sometimes infeasible target states can cause a high variance in the episode returns, which may make the value predictor network unreliable. We further regularize the training by only using actions with positive advantages \cite{van2007reinforcement,hamalainen2018ppo}.

\section{Movement Optimization using Low-Level Controllers} \label{section:movement_optimization}
The training procedure of the previous section produces a set of $H_{max}$ LLCs that can be used in movement optimization. This means the the action space becomes the space of target state trajectories of length $H\le H_{max}$. This action space is generic and can be used in offline/online trajectory optimization as well as reinforcement learning, as detailed below. 

\textbf{\textit{Offline Trajectory Optimization:}} In offline trajectory optimization settings, we optimize a trajectory of $4$ seconds using CMA-ES \cite{hansen2006cma}, i.e., every CMA-ES sample defines a trajectory of $T=40$ target states. In order to simulate each trajectory, at each timestep we use the next $H$ target states in the trajectory to query the LLC, which computes the low-level action. Similar to the other configurations, each low-level action is repeated for $0.1$ seconds. For the last $H-1$ target states, the target state sequences will be shorter than $H$ steps, and LLCs with smaller $H$ have to be used. 

\textbf{\textit{Online Trajectory Optimization:}} We use a simplified version of Fixed-Depth Informed MCTS (FDI-MCTS) \cite{Rajamaeki2018}, that performs a number of simulation rollouts from the current state using randomly sampled action trajectories of $2$ seconds (i.e., trajectory length $T=20$). The rollout actions are drawn randomly from a Gaussian distribution, with mean equal to the actions of the best trajectory of the previous timestep. For the last rollout action, the previous best trajectory is not available, and the action is sampled uniformly within the action space bounds. 

The exploration noise of the rollouts is proportional to the rollout index: $\boldsymbol{\sigma}^2=\frac{i+1}{N}(M-m)\textbf{I}$, where $i$ is the current rollout index, $N$ is the total number of rollouts, $m$ and $M$ are the actions' minimum and maximum values, and $\textbf{I}$ is the identity matrix. The rationale for this is that an online optimizer should both try to refine the current best solution and keep exploring with a large variance, in order to be able to adapt to sudden changes.

The rollouts are simulated forward from the current state in parallel, using multiple threads. After each timestep, badly performing rollouts are terminated, and the simulation resources are reassigned by forking a randomly chosen non-terminated rollout. In effect, the rollouts perform simple tree search. After simulating the rollouts up to the planning horizon, the first action of the best rollout is returned as the approximately optimal action. 

\textbf{\textit{Reinforcement Learning:}} For reinforcement learning, we use Proximal Policy Optimization (PPO) \cite{schulman2017proximal}, Soft Actor-Crit (SAC) \cite{haarnoja2018soft}, and Twin-Delayed Deep Deterministic Policy Gradient (TD3) \cite{fujimoto2018addressing}, three popular RL algorithms for continuous control.


\textbf{\textit{Implementation Details}} For the LLCs we used multi-layer perceptron (MLP) neural networks using Tensorflow \cite{tensorflow2015-whitepaper} and the Swish activation function \cite{ramachandran2017searching}. For \textit{HalfCheetah-v2}, \textit{Walked2d-v2}, and \textit{Hopper-v2}, the LLC policies had $2$ hidden layers of $64$ neurons, and for \textit{Humanoid-v2} we used larger policies with $3$ hidden layers of $128$ neurons. While building the exploration buffer and training the low-level controllers, we bypassed the default termination conditions defined in OpenAI Gym to make sure that the exploration is diverse and task-agnostic. The RL training uses PPO, SAC, and TD3 implementations from the \textit{stable-baselines} repository \cite{stable-baselines}. The hyperparameters used in the experiments are given in Appendix A.


\section{Experiments} \label{sec:experiments}
This section empirically compares the different action spaces discussed above. To reiterate, we compare the following:

\begin{itemize}
    \item \textit{Baseline}: The default action space without using low-level controllers, i.e., joint torques.
    \item \textit{LLC[Naive]}: The action space posed by the LLCs trained using the naive exploration approach with less diverse episode initialization.
    \item \textit{LLC[Contact-Based]}: The action space posed by the LLCs trained using our proposed contact-based exploration approach.
\end{itemize}

The LLCs are queried with a sequence of target states, $H$ states in total. For completeness, we also tested a version where a single target state was given to be reached $H$ steps in the future. However, as detailed in Appendix B, this results in overall worse performance.

The experiments below are designed to answer the following research questions:
\begin{itemize}
    \item \textit{RQ1}: Which action space works best, on average? Does using the LLC and training with the proposed contact-based exploration method improve optimization efficiency?
    
    \item \textit{RQ2}: How does the LLC planning horizon $H$ affect the optimization? Which $H$ should one use?
    
    \item \textit{RQ3}: How consistent are the results across the various tasks, agents, and optimization approaches?
    
    \item \textit{RQ4}: How does using the LLC affect the optimization landscape? Why does it make optimization easier?
\end{itemize} 
To ensure that the results are generalizable, the experiments are performed across the four agents and six tasks described in Section \ref{sec:simulation_environment} as well as the four different optimization approaches described in Section \ref{section:movement_optimization}.



\begin{figure*}[ht]
\centering
{{\includegraphics[width=1.0\linewidth]{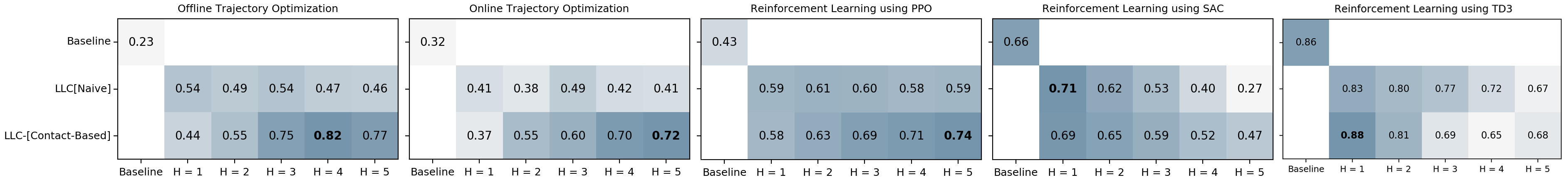} }}
\caption{Comparing different action spaces (baseline vs. LLC versions) and $H$ values in terms of normalized average episode/trajectory returns (higher is better) in the four optimization approaches. Overall, using an LLC outperforms the baseline, and LLCs trained using our contact-based exploration outperform LLCs trained using naive random exploration.}
\label{fig:comparing_hs}
\end{figure*}

\subsection{Comparing Action Spaces}\label{sec:rq1}

Fig. \ref{fig:comparing_hs} shows the benchmark results using normalized aggregate scores (higher is better) that measure the overall performance of each action space and optimization method across all agents and tasks. Using the LLC generally leads to significantly better results than the baseline, except with SAC and TD3, where the improvement is only marginal. Furthermore, LLCs trained with contact-based exploration in Algorithm \ref{alg:contact_exploration} mostly outperform the ones trained using naive exploration. 

To obtain the results, we performed 10 independent optimization runs with different random seeds for each combination of agent, task, optimizer, action space, and LLC horizon $H$. In total, this yields almost $20,000$ optimization runs. The simulation budget used in reinforcement learning using PPO, SAC, and TD3 is $10M$ simulation timesteps for \textit{Humanoid-v2} and $1M$  simulation timesteps for each of the other three environments. After all runs, the mean episode returns of the final iterations were normalized over each agent-task pair so that the worst run had score 0 and the best run had score 1. The scores in Fig. \ref{fig:comparing_hs} are averages of the normalized returns over all tasks and agents, i.e., a total of $10\times4\times6=240$ runs per cell, which should yield reasonably reliable results. 

\subsection{Effect of H}
RQ2 concerns the LLC horizon $H$, which is the most important tuning parameter in our approach and also determines the number of LLCs to be trained in Algorithm \ref{alg:llc_training}. 

As it can be seen in Fig. \ref{fig:comparing_hs}, three out of five approaches (offline/online trajectory optimization and PPO) work better with higher $H$ values, i.e., $H=4$ and $H=5$. In contrast, SAC and TD3 score highest with $H=1$ and their performance declines consistently when $H$ grows. We did not perform an exhaustive evaluation using $H>5$, as in our early experiments, further increasing $H$ did not improve the results with any of the optimization approaches. 

\subsection{Consistency Across Tasks and Agents} \label{sec:consistency}
The aggregate scores in Fig. \ref{fig:comparing_hs} do not allow comparing the consistency and variability of the results. Thus, to answer RQ3, we also scrutinized the performance of each agent and task. We simplified the investigation by only considering the best-performing $H$, i.e., $H=4$ for offline trajectory optimization, $H=5$ for online trajectory optimization and PPO, and $H=1$ for SAC and TD3.

\begin{figure*}[!ht]
\centering
{\includegraphics[width=1\linewidth]{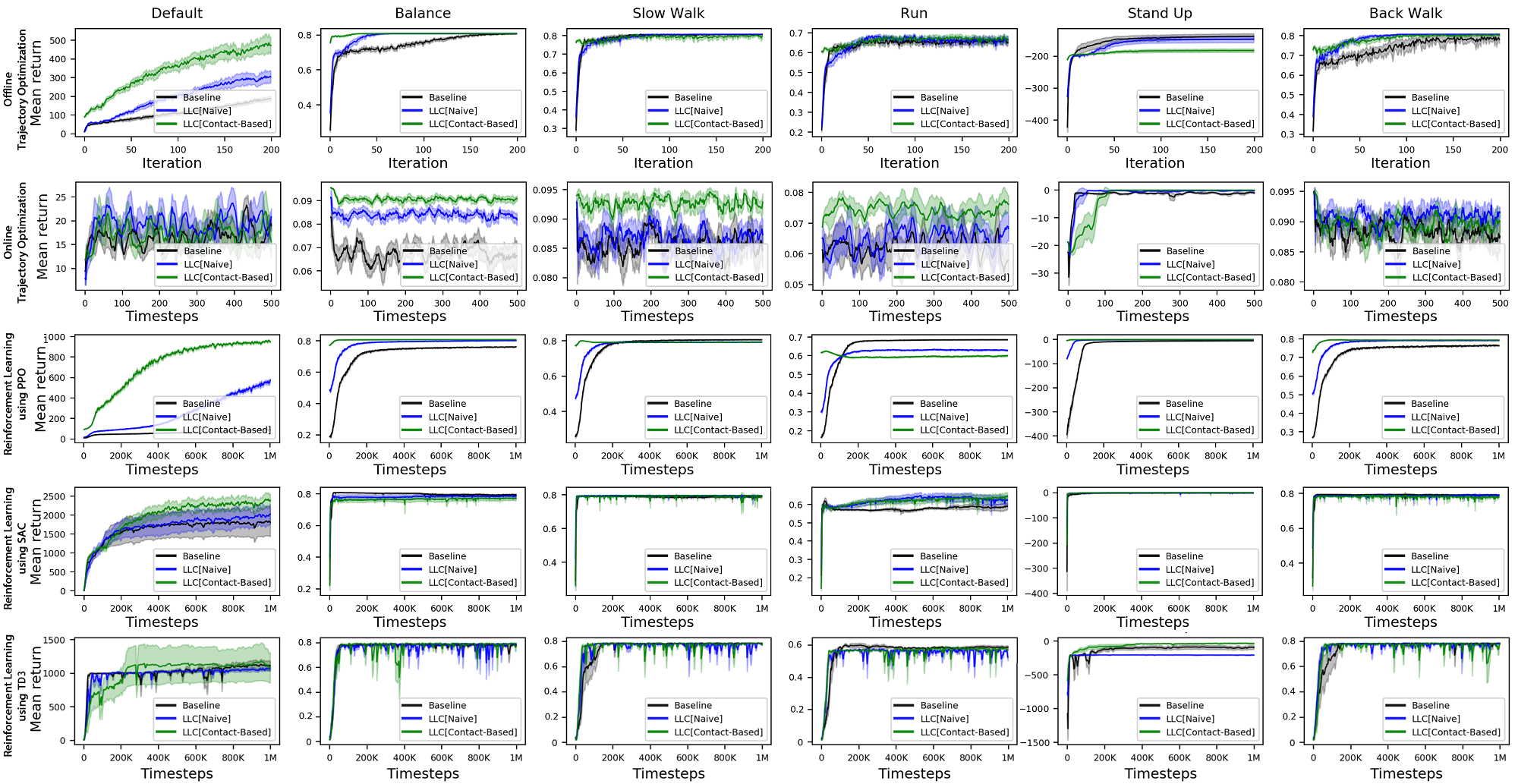} }
\caption{Results of offline/online trajectory optimization and reinforcement learning (using PPO , SAC, and TD3) in the Hopper-v2 environment. Each row corresponds to a different movement optimization approach, and each column corresponds to a different movement task.}
\label{fig:convergence_merged_hopper}
\end{figure*}

We outline the findings below, and Fig. \ref{fig:convergence_merged_hopper} shows example convergence plots from the Hopper-v2 environment. Full results of all environments can be found in Appendix C.

\textbf{\textit{Offline Trajectory Optimization:}} Both LLC versions are clearly better than the baseline in all tasks except for the stand up task. However, even in this case the final performance is comparable to the baseline. Overall, the contact-based exploration clearly surpasses the naive version.

\textbf{\textit{Online Trajectory Optimization:}} The results are consistent with those obtained in offline trajectory optimization: Both LLC-based approaches improve over the baseline, except in the stand up task. Again, the version with the contact-based exploration beats the one with naive exloration.

\textbf{\textit{Reinforcement Learning using PPO:}} Again, LLC with contact-based exploration performs best in all tasks, including the stand up task, but the difference is less pronounced with the humanoid agent. 

\textbf{\textit{Reinforcement Learning using SAC:}} In this configuration, all action spaces perform roughly similarly, except for a few tasks. LLC yields significant gains in the hopper default and humanoid stand-up tasks, but performs worse in the humanoid default task. 

\textbf{\textit{Reinforcement Learning using TD3:}} Similar to SAC, there are no major differences between the action spaces. 

The overall worse results of the LLC with the humanoid agent---while still providing significant gains with PPO and offline trajectory optimization---are likely due to both the higher state dimensionality and the aggressive episode termination whenever the agent starts to fall down. The high dimensionality makes the exploration data more sparse, and because the termination limits the agent to a small subregion of the state space, the contact-based diverse exploration can be expected to yield less gains. Fittingly, the LLC performs best with SAC in the stand up task which does not use termination.

To summarize the findings in Fig. \ref{fig:convergence_merged_hopper}, our approach is particularly effective in both offline and online trajectory optimization settings as well as reinforcement learning using PPO. 

\subsection{Effect of LLC on the Optimization Landscape}
So far, our results show a clear advantage of using the LLC, but the data is insufficient for analyzing the underlying reasons. A plausible explanation for trajectory optimization is that when not using the LLC and using control torques as actions, a small change to a single action can cause a large divergence in the rest of the trajectory, e.g., making a bipedal character fall. Furthermore, as pointed out in the case of a simple inverted pendulum in \cite{hamalainen2020visualizing}, the sum of rewards or costs over the trajectory is more sensitive to the first actions, which makes the optimization ill-conditioned, exhibiting elongated optima where an optimizer slowly zigzags between the cost function walls, or along a reward function ridge. In contrast, when using the LLC, perturbing the action of a single timestep---i.e., a single target state in a sequence---has only a small effect on states and rewards of the other timesteps.

\begin{figure*}[!t]
\centering
{\includegraphics[width=0.725\linewidth]{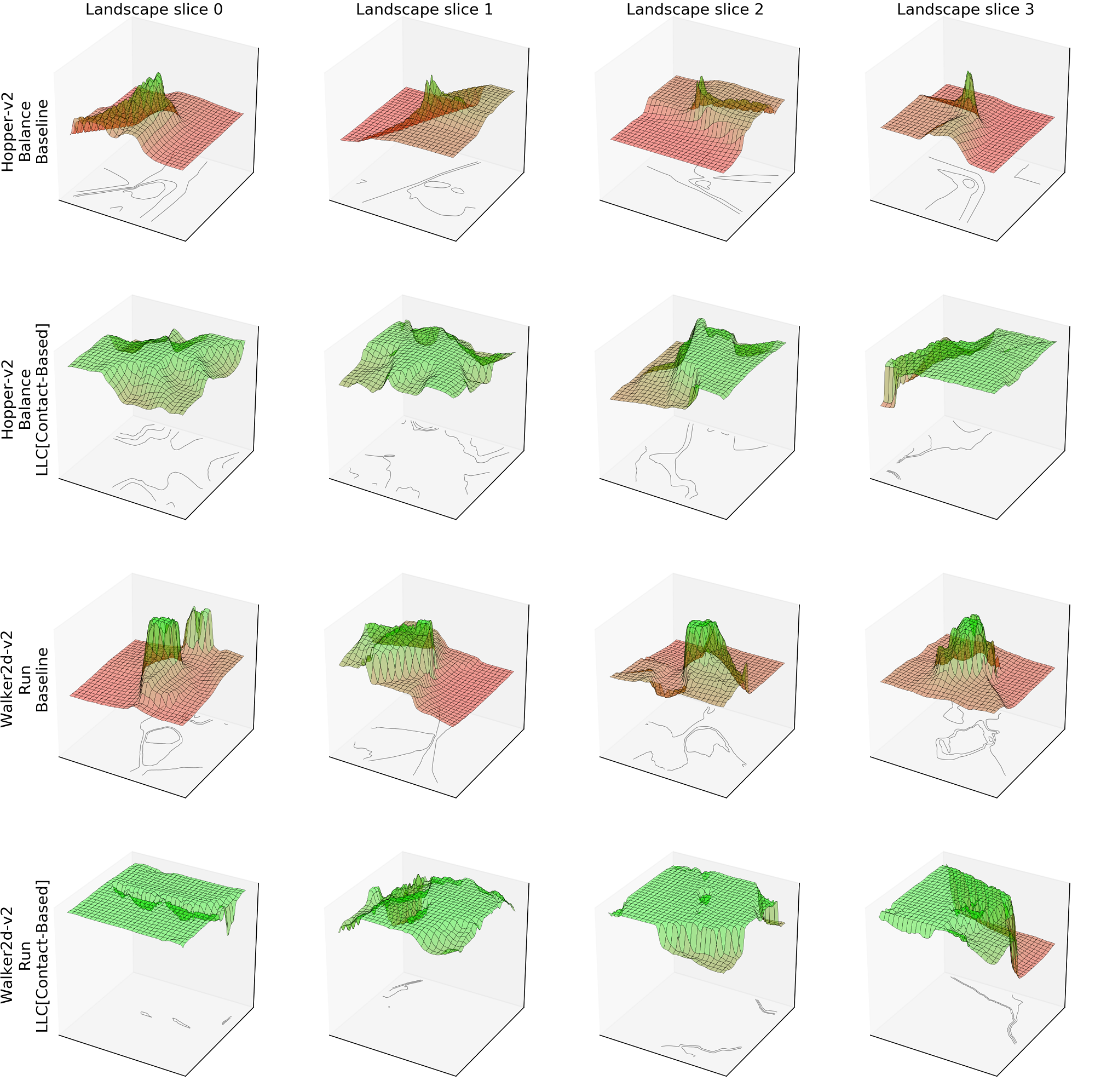}}
\caption{Offline trajectory optimization landscapes of Walker2d-v2 running and Hopper-v2 balancing with/without the low-level controllers (the up-axis shows the average episode return, i.e., higher is better). Using low-level controllers leads to smoother landscapes with significantly wider high-return basins that are easy to find using sampling-based optimization.}
\label{fig:landscape_plots_to}
\end{figure*}

In light of the above, optimizing without LLC should exhibit much narrower and/or elongated optima which slow down the convergence. This is a hypothesis we can test using the random 2D landscape slice visualization approach of \cite{li2018visualizing}, also applied to movement optimization by \cite{hamalainen2020visualizing}. This means that a multidimensional objective function is characterized by plotting it around the found optimum along a 2D slice of the parameter space defined by a pair of random orthogonal 2D subspace basis vectors. Fig. \ref{fig:landscape_plots_to} shows such 2D slices of offline trajectory optimization of the hopper balance and walker run tasks. As can be seen in the figure, using low-level controllers leads to smoother optimization landscapes with large and less ill-conditioned high-return basins. A similar visualization in the case of reinforcement learning can be found in Appendix D.

\section{Limitations and Future Work}
Considering the results, a clear limitation of our approach is that although we are able to improve the efficiency of trajectory optimization as well as on-policy RL using PPO, the improvements are only marginal in off-policy RL using SAC and TD3. When using SAC or TD3 and the best-performing choice for $H$, the compared action spaces yield approximately similar results in almost all the tasks. Future work is needed to investigate why this is the case. We hypothesize that RL algorithm differences in value estimation and learning play a role. In PPO, this is based on episode returns, which are more sensitive to small action perturbations when not using the LLC, similar to the trajectory returns of trajectory optimization. SAC and TD3’s Bellman backup using the twin Q networks might be less affected by the choice of action space.

In trajectory optimization, the stand up task did not benefit from using the LLC. One explanation for this is that the task requires more aggressive movement than the other tasks, using the LLC appears to bias the agent's initial movement exploration towards smooth and continuous movements. On the other hand, the problem does not persist in policy optimization. 


A further limitation is that we assume a full target observation is known for the LLC. Learning an HLC or directly using an LLC---e.g., for keyframe animation purposes---could be easier if one could also express the importance of each target observation variable. For example, one could produce locomotion by specifying a desired root rotation and velocity, and leave the other degrees of freedom to be decided by the LLC. We are currently investigating this, e.g., through the incorporation of recent machine learning models that can infer any number of unspecified variables  \cite{hamalainen2020deep}. Another potential topic for future work could be to use our contact-based exploration method in offline RL methods such as conservative Q-learning (CQL) \cite{kumar2020conservative}. The added stability provided by an LLC could make it easier to learn from limited data.


Finally, future work is needed to extend the initial state randomization of our contact-based exploration algorithm to object manipulation tasks, and improve the robustness of the LLCs. In the current version, the environment is assumed to be static, which is not valid for, e.g., real-world robotic manipulation tasks. Our LLCs are also not robust enough for following any target trajectory without errors, and the HLC has to compensate for the LLC's imperfections. 

\section{Conclusion}
We have proposed a hierarchical movement control approach where a low-level controller (LLC) is trained to follow target state trajectories, and movement optimization---including both trajectory optimization and reinforcement learning---can then operate in the space of target states instead of low-level actions such as control torques. Through extensive experiments across multiple agents, tasks, and optimization methods, we have shown that our LLCs improve movement optimization convergence and the attainable objective function values. This can be explained by the LLC making state-space movement trajectories less sensitive to small changes of the actions, which results in wider optima that are easier to find. 

In contrast to previous work that trains LLCs using motion capture data and is thus limited in terms of supported agents and movements, we have proposed a simple contact-based exploration method to synthesize diverse, task-agnostic LLC training data for a variety of agents that depend on contact forces for actuation, e.g., bipeds and quadrupeds. Our experiments also indicate that LLCs trained with our exploration data perform better than LLCs trained with baseline random exploration data. We believe our work provides a building block towards a general solution to the movement optimization problem, improving a wide range of movement optimization applications.



\ifCLASSOPTIONcompsoc
  \section*{Acknowledgments}
\else
  \section*{Acknowledgment}
\fi

This work was supported by Academy of Finland grant 299358 and the Technology Industries of Finland Centennial Foundation. The experiments utilized the Triton cloud computing infrastructure of Aalto University. Part of the research was conducted while the first author was a visiting researcher at University of British Columbia, Canada.

\ifCLASSOPTIONcaptionsoff
  \newpage
\fi



\bibliographystyle{IEEEtran}
\bibliography{references}

\begin{thebibliography}{10}
\providecommand{\url}[1]{#1}
\csname url@samestyle\endcsname
\providecommand{\newblock}{\relax}
\providecommand{\bibinfo}[2]{#2}
\providecommand{\BIBentrySTDinterwordspacing}{\spaceskip=0pt\relax}
\providecommand{\BIBentryALTinterwordstretchfactor}{4}
\providecommand{\BIBentryALTinterwordspacing}{\spaceskip=\fontdimen2\font plus
\BIBentryALTinterwordstretchfactor\fontdimen3\font minus
  \fontdimen4\font\relax}
\providecommand{\BIBforeignlanguage}[2]{{%
\expandafter\ifx\csname l@#1\endcsname\relax
\typeout{** WARNING: IEEEtran.bst: No hyphenation pattern has been}%
\typeout{** loaded for the language `#1'. Using the pattern for}%
\typeout{** the default language instead.}%
\else
\language=\csname l@#1\endcsname
\fi
#2}}
\providecommand{\BIBdecl}{\relax}
\BIBdecl

\bibitem{tassa2012synthesis}
Y.~Tassa, T.~Erez, and E.~Todorov, ``Synthesis and stabilization of complex
  behaviors through online trajectory optimization,'' in \emph{Intelligent
  Robots and Systems (IROS), 2012 IEEE/RSJ International Conference on}.\hskip
  1em plus 0.5em minus 0.4em\relax IEEE, 2012, pp. 4906--4913.

\bibitem{naderi2017discovering}
\BIBentryALTinterwordspacing
K.~Naderi, J.~Rajam\"{a}ki, and P.~H\"{a}m\"{a}l\"{a}inen, ``Discovering and
  synthesizing humanoid climbing movements,'' \emph{ACM Transactions on
  Graphics (TOG)}, vol.~36, no.~4, pp. 43:1--43:11, Jul. 2017. [Online].
  Available: \url{http://doi.acm.org/10.1145/3072959.3073707}
\BIBentrySTDinterwordspacing

\bibitem{bergamin2019drecon}
K.~Bergamin, S.~Clavet, D.~Holden, and J.~R. Forbes, ``Drecon: data-driven
  responsive control of physics-based characters,'' \emph{ACM Transactions On
  Graphics (TOG)}, vol.~38, no.~6, pp. 1--11, 2019.

\bibitem{2018-TOG-deepMimic}
\BIBentryALTinterwordspacing
X.~B. Peng, P.~Abbeel, S.~Levine, and M.~van~de Panne, ``{DeepMimic:
  Example-guided deep reinforcement learning of physics-based character
  skills},'' \emph{ACM Transactions on Graphics (TOG)}, vol.~37, no.~4, pp.
  143:1--143:14, Jul. 2018. [Online]. Available:
  \url{http://doi.acm.org/10.1145/3197517.3201311}
\BIBentrySTDinterwordspacing

\bibitem{won2020scalable}
J.~Won, D.~Gopinath, and J.~Hodgins, ``A scalable approach to control diverse
  behaviors for physically simulated characters,'' \emph{ACM Transactions on
  Graphics (TOG)}, vol.~39, no.~4, pp. 33--1, 2020.

\bibitem{peng2017learning}
X.~B. Peng and M.~van~de Panne, ``Learning locomotion skills using deeprl: Does
  the choice of action space matter?'' in \emph{Proceedings of the ACM
  SIGGRAPH/Eurographics Symposium on Computer Animation}, 2017, pp. 1--13.

\bibitem{Merel2020Deep}
\BIBentryALTinterwordspacing
J.~Merel, D.~Aldarondo, J.~Marshall, Y.~Tassa, G.~Wayne, and B.~Olveczky,
  ``Deep neuroethology of a virtual rodent,'' in \emph{International Conference
  on Learning Representations}, 2020. [Online]. Available:
  \url{https://openreview.net/forum?id=SyxrxR4KPS}
\BIBentrySTDinterwordspacing

\bibitem{peng2017deeploco}
X.~B. Peng, G.~Berseth, K.~Yin, and M.~Van De~Panne, ``Deeploco: dynamic
  locomotion skills using hierarchical deep reinforcement learning,'' \emph{ACM
  Transactions on Graphics (TOG)}, vol.~36, no.~4, p.~41, 2017.

\bibitem{levy2019learning}
A.~Levy, G.~Konidaris, R.~Platt, and K.~Saenko, ``Learning multi-level
  hierarchies with hindsight,'' in \emph{Proceedings of International
  Conference on Learning Representations}, 2019.

\bibitem{lee2019scalable}
S.~Lee, M.~Park, K.~Lee, and J.~Lee, ``Scalable muscle-actuated human
  simulation and control,'' \emph{ACM Transactions On Graphics (TOG)}, vol.~38,
  no.~4, pp. 1--13, 2019.

\bibitem{hamalainen2020visualizing}
P.~H{\"a}m{\"a}l{\"a}inen, J.~Toikka, A.~Babadi, and K.~Liu, ``Visualizing
  movement control optimization landscapes,'' \emph{IEEE Transactions on
  Visualization and Computer Graphics}, August 2020.

\bibitem{yin2007simbicon}
K.~Yin, K.~Loken, and M.~Van~de Panne, ``Simbicon: Simple biped locomotion
  control,'' \emph{ACM Transactions on Graphics (TOG)}, vol.~26, no.~3, pp.
  105--es, 2007.

\bibitem{Yin08}
K.~Yin, S.~Coros, P.~Beaudoin, and M.~van~de Panne, ``Continuation methods for
  adapting simulated skills,'' \emph{ACM Trans. Graph.}, vol.~27, no.~3, 2008.

\bibitem{Geijtenbeek2013}
T.~Geijtenbeek, M.~van~de Panne, and A.~F. van~der Stappen, ``Flexible
  muscle-based locomotion for bipedal creatures,'' \emph{ACM Transactions on
  Graphics (TOG)}, vol.~32, no.~6, 2013.

\bibitem{hansen2006cma}
N.~Hansen, ``The cma evolution strategy: a comparing review,'' in \emph{Towards
  a new evolutionary computation}.\hskip 1em plus 0.5em minus 0.4em\relax
  Springer, 2006, pp. 75--102.

\bibitem{mordatch2012discovery}
I.~Mordatch, E.~Todorov, and Z.~Popovi{\'c}, ``Discovery of complex behaviors
  through contact-invariant optimization,'' \emph{ACM Transactions on Graphics
  (TOG)}, vol.~31, no.~4, p.~43, 2012.

\bibitem{naderi2018learning}
K.~Naderi, A.~Babadi, and P.~H{\"a}m{\"a}l{\"a}inen, ``Learning physically
  based humanoid climbing movements,'' in \emph{Computer Graphics Forum},
  vol.~37, no.~8.\hskip 1em plus 0.5em minus 0.4em\relax Wiley Online Library,
  2018, pp. 69--80.

\bibitem{hamalainen2014online}
P.~H{\"a}m{\"a}l{\"a}inen, S.~Eriksson, E.~Tanskanen, V.~Kyrki, and
  J.~Lehtinen, ``Online motion synthesis using sequential monte carlo,''
  \emph{ACM Transactions on Graphics (TOG)}, vol.~33, no.~4, p.~51, 2014.

\bibitem{babadi2018intelligent}
A.~Babadi, K.~Naderi, and P.~H{\"a}m{\"a}l{\"a}inen, ``Intelligent middle-level
  game control,'' in \emph{Proceedings of IEEE Conference on Computational
  Intelligence and Games (IEEE CIG)}.\hskip 1em plus 0.5em minus 0.4em\relax
  IEEE, 2018.

\bibitem{dayan1993feudal}
P.~Dayan and G.~E. Hinton, ``Feudal reinforcement learning,'' in \emph{Advances
  in neural information processing systems}, 1993, pp. 271--278.

\bibitem{sutton1999between}
R.~S. Sutton, D.~Precup, and S.~Singh, ``Between mdps and semi-mdps: A
  framework for temporal abstraction in reinforcement learning,''
  \emph{Artificial intelligence}, vol. 112, no. 1-2, pp. 181--211, 1999.

\bibitem{dietterich2000hierarchical}
T.~G. Dietterich, ``Hierarchical reinforcement learning with the maxq value
  function decomposition,'' \emph{Journal of artificial intelligence research},
  vol.~13, pp. 227--303, 2000.

\bibitem{2015-TOG-terrainRL}
\BIBentryALTinterwordspacing
X.~B. Peng, G.~Berseth, and M.~van~de Panne, ``Dynamic terrain traversal skills
  using reinforcement learning,'' \emph{ACM Transactions on Graphics (TOG)},
  vol.~34, no.~4, pp. 80:1--80:11, Jul. 2015. [Online]. Available:
  \url{http://doi.acm.org/10.1145/2766910}
\BIBentrySTDinterwordspacing

\bibitem{2016-TOG-deepRL}
------, ``Terrain-adaptive locomotion skills using deep reinforcement
  learning,'' \emph{ACM Transactions on Graphics (Proc. SIGGRAPH 2016)},
  vol.~35, no.~4, 2016.

\bibitem{liu2017learning}
L.~Liu and J.~Hodgins, ``Learning to schedule control fragments for
  physics-based characters using deep q-learning,'' \emph{ACM Transactions on
  Graphics (TOG)}, vol.~36, no.~3, pp. 1--14, 2017.

\bibitem{mnih2015human}
V.~Mnih, K.~Kavukcuoglu, D.~Silver, A.~A. Rusu, J.~Veness, M.~G. Bellemare,
  A.~Graves, M.~Riedmiller, A.~K. Fidjeland, G.~Ostrovski \emph{et~al.},
  ``Human-level control through deep reinforcement learning,'' \emph{Nature},
  vol. 518, no. 7540, p. 529, 2015.

\bibitem{MCPPeng19}
X.~B. Peng, M.~Chang, G.~Zhang, P.~Abbeel, and S.~Levine, ``Mcp: Learning
  composable hierarchical control with multiplicative compositional policies,''
  in \emph{Advances in Neural Information Processing Systems 32}.\hskip 1em
  plus 0.5em minus 0.4em\relax Curran Associates, Inc., 2019, pp. 3681--3692.

\bibitem{nachum2018near}
O.~Nachum, S.~Gu, H.~Lee, and S.~Levine, ``Near-optimal representation learning
  for hierarchical reinforcement learning,'' \emph{arXiv preprint
  arXiv:1810.01257}, 2018.

\bibitem{andrychowicz2017hindsight}
M.~Andrychowicz, F.~Wolski, A.~Ray, J.~Schneider, R.~Fong, P.~Welinder,
  B.~McGrew, J.~Tobin, O.~P. Abbeel, and W.~Zaremba, ``Hindsight experience
  replay,'' in \emph{Advances in neural information processing systems}, 2017,
  pp. 5048--5058.

\bibitem{nachum2018data}
O.~Nachum, S.~S. Gu, H.~Lee, and S.~Levine, ``Data-efficient hierarchical
  reinforcement learning,'' in \emph{Advances in Neural Information Processing
  Systems}, 2018, pp. 3303--3313.

\bibitem{merel2017learning}
J.~Merel, Y.~Tassa, D.~TB, S.~Srinivasan, J.~Lemmon, Z.~Wang, G.~Wayne, and
  N.~Heess, ``Learning human behaviors from motion capture by adversarial
  imitation,'' \emph{arXiv preprint arXiv:1707.02201}, 2017.

\bibitem{luo2020carl}
Y.-S. Luo, J.~H. Soeseno, T.~P.-C. Chen, and W.-C. Chen, ``Carl: Controllable
  agent with reinforcement learning for quadruped locomotion,'' \emph{ACM
  Transactions on Graphics (TOG)}, vol.~39, no.~4, pp. 38--1, 2020.

\bibitem{merel2018hierarchical}
J.~Merel, A.~Ahuja, V.~Pham, S.~Tunyasuvunakool, S.~Liu, D.~Tirumala, N.~Heess,
  and G.~Wayne, ``Hierarchical visuomotor control of humanoids,'' \emph{arXiv
  preprint arXiv:1811.09656}, 2018.

\bibitem{merel2020catch}
J.~Merel, S.~Tunyasuvunakool, A.~Ahuja, Y.~Tassa, L.~Hasenclever, V.~Pham,
  T.~Erez, G.~Wayne, and N.~Heess, ``Catch \& carry: reusable neural
  controllers for vision-guided whole-body tasks,'' \emph{ACM Transactions on
  Graphics (TOG)}, vol.~39, no.~4, pp. 39--1, 2020.

\bibitem{merel2018neural}
J.~Merel, L.~Hasenclever, A.~Galashov, A.~Ahuja, V.~Pham, G.~Wayne, Y.~W. Teh,
  and N.~Heess, ``Neural probabilistic motor primitives for humanoid control,''
  \emph{arXiv preprint arXiv:1811.11711}, 2018.

\bibitem{hasenclever2020comic}
L.~Hasenclever, F.~Pardo, R.~Hadsell, N.~Heess, and J.~Merel, ``Comic:
  Complementary task learning \& mimicry for reusable skills,'' in
  \emph{International Conference on Machine Learning}.\hskip 1em plus 0.5em
  minus 0.4em\relax PMLR, 2020, pp. 4105--4115.

\bibitem{Park:2019}
S.~Park, H.~Ryu, S.~Lee, S.~Lee, and J.~Lee, ``Learning predict-and-simulate
  policies from unorganized human motion data,'' \emph{ACM Trans. Graph.},
  vol.~38, no.~6, 2019.

\bibitem{levine2013guided}
S.~Levine and V.~Koltun, ``Guided policy search,'' in \emph{International
  conference on machine learning}.\hskip 1em plus 0.5em minus 0.4em\relax PMLR,
  2013, pp. 1--9.

\bibitem{mordatch2014combining}
I.~Mordatch and E.~Todorov, ``Combining the benefits of function approximation
  and trajectory optimization.'' in \emph{Robotics: Science and Systems},
  vol.~4, 2014.

\bibitem{rajamaki2017augmenting}
J.~Rajam{\"a}ki and P.~H{\"a}m{\"a}l{\"a}inen, ``Augmenting sampling based
  controllers with machine learning,'' in \emph{Proceedings of the ACM
  SIGGRAPH/Eurographics Symposium on Computer Animation}.\hskip 1em plus 0.5em
  minus 0.4em\relax ACM, 2017, p.~11.

\bibitem{won2017train}
J.~Won, J.~Park, K.~Kim, and J.~Lee, ``How to train your dragon: example-guided
  control of flapping flight,'' \emph{ACM Transactions on Graphics (TOG)},
  vol.~36, no.~6, pp. 1--13, 2017.

\bibitem{babadi2019self}
A.~Babadi, K.~Naderi, and P.~H{\"a}m{\"a}l{\"a}inen, ``Self-imitation learning
  of locomotion movements through termination curriculum,'' in \emph{Motion,
  Interaction and Games}, 2019, pp. 1--7.

\bibitem{thrun2002probabilistic}
S.~Thrun, ``Probabilistic robotics,'' \emph{Communications of the ACM},
  vol.~45, no.~3, pp. 52--57, 2002.

\bibitem{Kaiser2020Model}
\BIBentryALTinterwordspacing
Łukasz Kaiser, M.~Babaeizadeh, P.~Miłos, B.~Osiński, R.~H. Campbell,
  K.~Czechowski, D.~Erhan, C.~Finn, P.~Kozakowski, S.~Levine, A.~Mohiuddin,
  R.~Sepassi, G.~Tucker, and H.~Michalewski, ``Model based reinforcement
  learning for atari,'' in \emph{International Conference on Learning
  Representations}, 2020. [Online]. Available:
  \url{https://openreview.net/forum?id=S1xCPJHtDB}
\BIBentrySTDinterwordspacing

\bibitem{xie2016model}
C.~Xie, S.~Patil, T.~Moldovan, S.~Levine, and P.~Abbeel, ``Model-based
  reinforcement learning with parametrized physical models and optimism-driven
  exploration,'' in \emph{2016 IEEE international conference on robotics and
  automation (ICRA)}.\hskip 1em plus 0.5em minus 0.4em\relax IEEE, 2016, pp.
  504--511.

\bibitem{boney2019regularizing}
R.~Boney, J.~Kannala, and A.~Ilin, ``Regularizing model-based planning with
  energy-based models,'' in \emph{2019 Conference on Robot Learning (CoRL)},
  2019.

\bibitem{orseau2013universal}
L.~Orseau, T.~Lattimore, and M.~Hutter, ``Universal knowledge-seeking agents
  for stochastic environments,'' in \emph{International Conference on
  Algorithmic Learning Theory}.\hskip 1em plus 0.5em minus 0.4em\relax
  Springer, 2013, pp. 158--172.

\bibitem{hester2017intrinsically}
T.~Hester and P.~Stone, ``Intrinsically motivated model learning for developing
  curious robots,'' \emph{Artificial Intelligence}, vol. 247, pp. 170--186,
  2017.

\bibitem{bellemare2016unifying}
M.~Bellemare, S.~Srinivasan, G.~Ostrovski, T.~Schaul, D.~Saxton, and R.~Munos,
  ``Unifying count-based exploration and intrinsic motivation,'' in
  \emph{Advances in neural information processing systems}, 2016, pp.
  1471--1479.

\bibitem{hafner2019dream}
D.~Hafner, T.~Lillicrap, J.~Ba, and M.~Norouzi, ``Dream to control: Learning
  behaviors by latent imagination,'' \emph{arXiv preprint arXiv:1912.01603},
  2019.

\bibitem{sekar2020planning}
R.~Sekar, O.~Rybkin, K.~Daniilidis, P.~Abbeel, D.~Hafner, and D.~Pathak,
  ``Planning to explore via self-supervised world models,'' \emph{arXiv
  preprint arXiv:2005.05960}, 2020.

\bibitem{liu2012terrain}
L.~Liu, K.~Yin, M.~van~de Panne, and B.~Guo, ``Terrain runner: control,
  parameterization, composition, and planning for highly dynamic motions.''
  \emph{ACM Trans. Graph.}, vol.~31, no.~6, pp. 154--1, 2012.

\bibitem{Rajamaeki2018}
J.~J. Rajam{\"a}ki and P.~H{\"a}m{\"a}l{\"a}inen, ``Continuous control monte
  carlo tree search informed by multiple experts,'' \emph{IEEE transactions on
  visualization and computer graphics}, 2018.

\bibitem{sutton2018reinforcement}
R.~S. Sutton and A.~G. Barto, \emph{Reinforcement learning: An
  introduction}.\hskip 1em plus 0.5em minus 0.4em\relax MIT press, 2018.

\bibitem{schulman2017proximal}
J.~Schulman, F.~Wolski, P.~Dhariwal, A.~Radford, and O.~Klimov, ``Proximal
  policy optimization algorithms,'' \emph{arXiv preprint arXiv:1707.06347},
  2017.

\bibitem{haarnoja2018soft}
T.~Haarnoja, A.~Zhou, P.~Abbeel, and S.~Levine, ``Soft actor-critic: Off-policy
  maximum entropy deep reinforcement learning with a stochastic actor,''
  \emph{arXiv preprint arXiv:1801.01290}, 2018.

\bibitem{fujimoto2018addressing}
S.~Fujimoto, H.~Hoof, and D.~Meger, ``Addressing function approximation error
  in actor-critic methods,'' in \emph{International Conference on Machine
  Learning}.\hskip 1em plus 0.5em minus 0.4em\relax PMLR, 2018, pp. 1587--1596.

\bibitem{lillicrap2015continuous}
T.~P. Lillicrap, J.~J. Hunt, A.~Pritzel, N.~Heess, T.~Erez, Y.~Tassa,
  D.~Silver, and D.~Wierstra, ``Continuous control with deep reinforcement
  learning,'' \emph{arXiv preprint arXiv:1509.02971}, 2015.

\bibitem{peng2020learning}
X.~B. Peng, E.~Coumans, T.~Zhang, T.-W. Lee, J.~Tan, and S.~Levine, ``Learning
  agile robotic locomotion skills by imitating animals,'' \emph{arXiv preprint
  arXiv:2004.00784}, 2020.

\bibitem{baselines}
P.~Dhariwal, C.~Hesse, O.~Klimov, A.~Nichol, M.~Plappert, A.~Radford,
  J.~Schulman, S.~Sidor, Y.~Wu, and P.~Zhokhov, ``Openai baselines,''
  \url{https://github.com/openai/baselines}, 2017.

\bibitem{hafner2017tensorflow}
D.~Hafner, J.~Davidson, and V.~Vanhoucke, ``Tensorflow agents: Efficient
  batched reinforcement learning in tensorflow,'' \emph{arXiv preprint
  arXiv:1709.02878}, 2017.

\bibitem{juliani2018unity}
A.~Juliani, V.-P. Berges, E.~Vckay, Y.~Gao, H.~Henry, M.~Mattar, and D.~Lange,
  ``Unity: A general platform for intelligent agents,'' \emph{arXiv preprint
  arXiv:1809.02627}, 2018.

\bibitem{pardo2020tonic}
F.~Pardo, ``Tonic: A deep reinforcement learning library for fast prototyping
  and benchmarking,'' \emph{arXiv preprint arXiv:2011.07537}, 2020.

\bibitem{schulman2015high}
J.~Schulman, P.~Moritz, S.~Levine, M.~Jordan, and P.~Abbeel, ``High-dimensional
  continuous control using generalized advantage estimation,'' \emph{arXiv
  preprint arXiv:1506.02438}, 2015.

\bibitem{todorov2012mujoco}
E.~Todorov, T.~Erez, and Y.~Tassa, ``Mujoco: A physics engine for model-based
  control,'' in \emph{2012 IEEE/RSJ International Conference on Intelligent
  Robots and Systems}.\hskip 1em plus 0.5em minus 0.4em\relax IEEE, 2012, pp.
  5026--5033.

\bibitem{brockman2016openai}
G.~Brockman, V.~Cheung, L.~Pettersson, J.~Schneider, J.~Schulman, J.~Tang, and
  W.~Zaremba, ``Openai gym,'' \emph{arXiv preprint arXiv:1606.01540}, 2016.

\bibitem{heess2017emergence}
N.~Heess, D.~TB, S.~Sriram, J.~Lemmon, J.~Merel, G.~Wayne, Y.~Tassa, T.~Erez,
  Z.~Wang, S.~Eslami \emph{et~al.}, ``Emergence of locomotion behaviours in
  rich environments,'' \emph{arXiv preprint arXiv:1707.02286}, 2017.

\bibitem{gupta2018unsupervised}
A.~Gupta, B.~Eysenbach, C.~Finn, and S.~Levine, ``Unsupervised meta-learning
  for reinforcement learning,'' \emph{arXiv preprint arXiv:1806.04640}, 2018.

\bibitem{chua2018deep}
K.~Chua, R.~Calandra, R.~McAllister, and S.~Levine, ``Deep reinforcement
  learning in a handful of trials using probabilistic dynamics models,'' in
  \emph{Advances in Neural Information Processing Systems}, 2018, pp.
  4754--4765.

\bibitem{ghosh2019learning}
D.~Ghosh, A.~Gupta, A.~Reddy, J.~Fu, C.~Devin, B.~Eysenbach, and S.~Levine,
  ``Learning to reach goals via iterated supervised learning,'' 2019.

\bibitem{schmeckpeper2019learning}
K.~Schmeckpeper, A.~Xie, O.~Rybkin, S.~Tian, K.~Daniilidis, S.~Levine, and
  C.~Finn, ``Learning predictive models from observation and interaction,''
  2019.

\bibitem{van2007reinforcement}
H.~Van~Hasselt and M.~A. Wiering, ``Reinforcement learning in continuous action
  spaces,'' in \emph{2007 IEEE International Symposium on Approximate Dynamic
  Programming and Reinforcement Learning}.\hskip 1em plus 0.5em minus
  0.4em\relax IEEE, 2007, pp. 272--279.

\bibitem{hamalainen2018ppo}
P.~H{\"a}m{\"a}l{\"a}inen, A.~Babadi, X.~Ma, and J.~Lehtinen,
  ``{P}{P}{O}-{C}{M}{A}: Proximal policy optimization with covariance matrix
  adaptation,'' \emph{arXiv preprint arXiv:1810.02541}, 2018.

\bibitem{tensorflow2015-whitepaper}
\BIBentryALTinterwordspacing
M.~Abadi, A.~Agarwal, P.~Barham, E.~Brevdo, Z.~Chen, C.~Citro, G.~S. Corrado,
  A.~Davis, J.~Dean, M.~Devin, S.~Ghemawat, I.~Goodfellow, A.~Harp, G.~Irving,
  M.~Isard, Y.~Jia, R.~Jozefowicz, L.~Kaiser, M.~Kudlur, J.~Levenberg,
  D.~Man\'{e}, R.~Monga, S.~Moore, D.~Murray, C.~Olah, M.~Schuster, J.~Shlens,
  B.~Steiner, I.~Sutskever, K.~Talwar, P.~Tucker, V.~Vanhoucke, V.~Vasudevan,
  F.~Vi\'{e}gas, O.~Vinyals, P.~Warden, M.~Wattenberg, M.~Wicke, Y.~Yu, and
  X.~Zheng, ``{TensorFlow}: Large-scale machine learning on heterogeneous
  systems,'' 2015, software available from tensorflow.org. [Online]. Available:
  \url{https://www.tensorflow.org/}
\BIBentrySTDinterwordspacing

\bibitem{ramachandran2017searching}
P.~Ramachandran, B.~Zoph, and Q.~V. Le, ``Searching for activation functions,''
  \emph{arXiv preprint arXiv:1710.05941}, 2017.

\bibitem{stable-baselines}
A.~Hill, A.~Raffin, M.~Ernestus, A.~Gleave, A.~Kanervisto, R.~Traore,
  P.~Dhariwal, C.~Hesse, O.~Klimov, A.~Nichol, M.~Plappert, A.~Radford,
  J.~Schulman, S.~Sidor, and Y.~Wu, ``Stable baselines,''
  \url{https://github.com/hill-a/stable-baselines}, 2018.

\bibitem{li2018visualizing}
H.~Li, Z.~Xu, G.~Taylor, C.~Studer, and T.~Goldstein, ``Visualizing the loss
  landscape of neural nets,'' in \emph{Advances in Neural Information
  Processing Systems}, 2018, pp. 6389--6399.

\bibitem{hamalainen2020deep}
P.~H{\"a}m{\"a}l{\"a}inen and A.~Solin, ``Deep residual mixture models,''
  \emph{arXiv preprint arXiv:2006.12063}, 2020.

\bibitem{kumar2020conservative}
A.~Kumar, A.~Zhou, G.~Tucker, and S.~Levine, ``Conservative q-learning for
  offline reinforcement learning,'' 2020.

\end{thebibliography}


\begin{IEEEbiography}[{\includegraphics[width=1in,height=1.25in,clip,keepaspectratio]{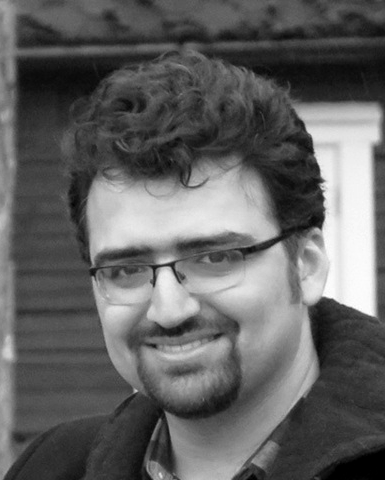}}]{Amin Babadi}
 is a doctoral candidate at the Department of Computer Science, Aalto University, Finland. His research focuses on developing efficient, creative movement artificial intelligence for physically simulated characters in multi-agent settings. Babadi has previously worked on three commercial games, developing AI, animation, gameplay, and physics simulation systems.
\end{IEEEbiography}

\begin{IEEEbiography}[{\includegraphics[width=1in,height=1.25in,clip,keepaspectratio]{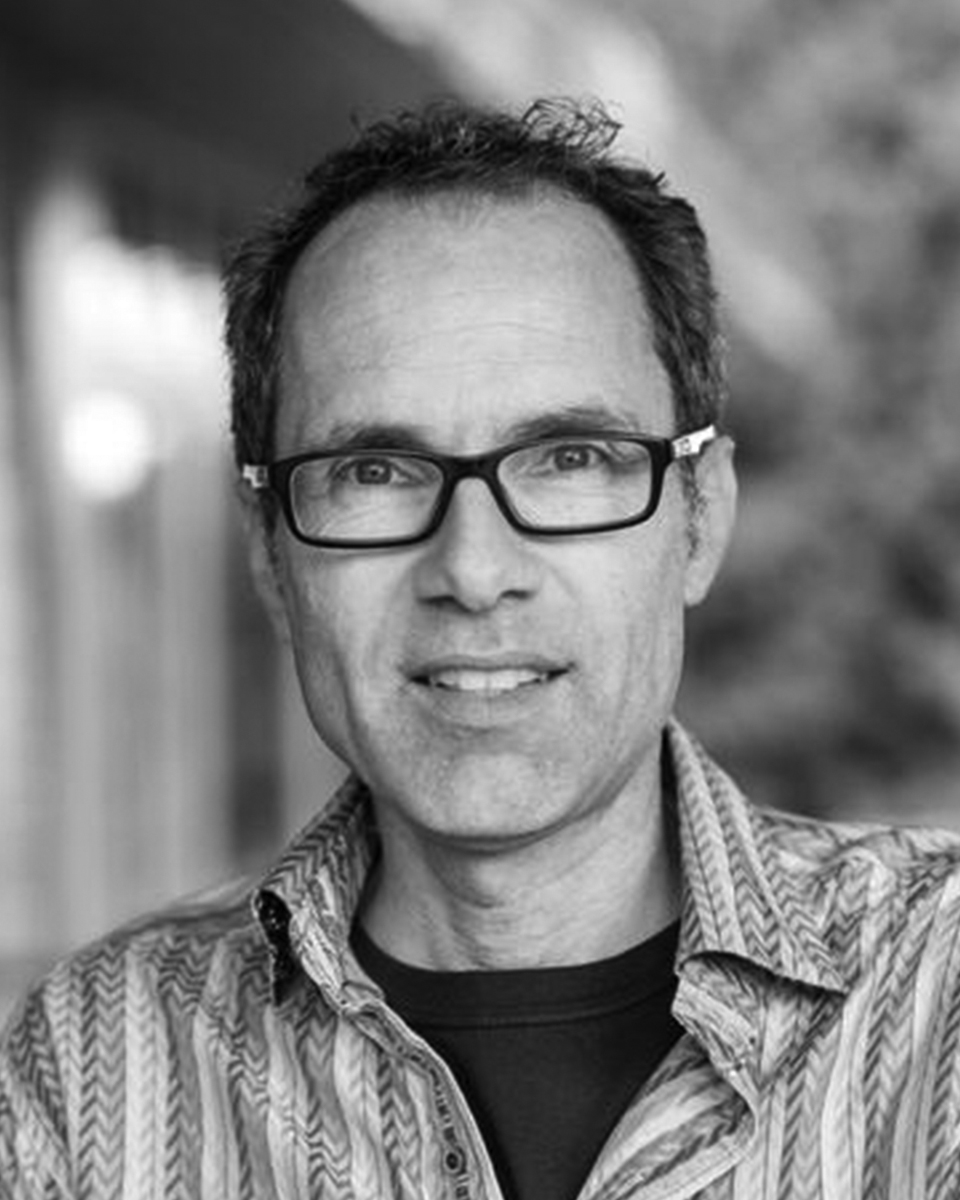}}]{Michiel van de Panne}
  obtained his B.A.Sc. in 1987 (University of Calgary), and his M.A.Sc. and Ph.D. in 1989 and 1994, respectively (University of Toronto). From 1993 to 2001 he was a faculty member in the Department of Computer Science at the University of Toronto. Since 2002 he has been with the Department of Computer Science at the University of British Columbia as Associate Professor (2002-2008) and as Full Professor (2008-). He served as Associate Head for Research and Faculty Affairs during 2011-14. During 2000-2001, he was a visiting professor at the University of British Columbia, and founded Motion Playground Inc. to develop games and educational applications using physics-based animation and simulation. He was a visiting Researcher at INRIA Sophia Antipolis during 2007-8.
\end{IEEEbiography}

\begin{IEEEbiography}[{\includegraphics[width=1in,height=1.25in,clip,keepaspectratio]{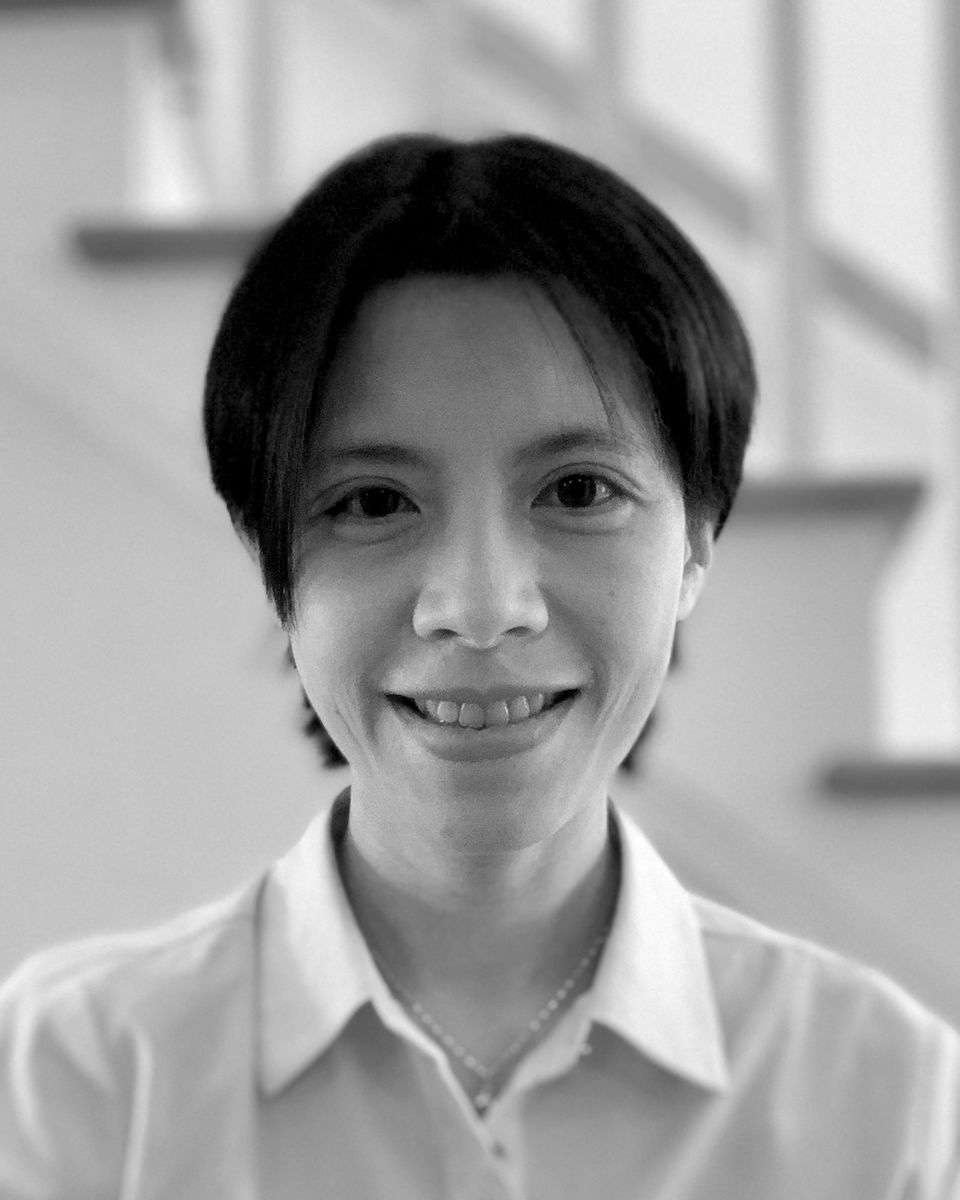}}]{C. Karen Liu}
  is an associate professor in the Department of Computer Science at Stanford University. She received her Ph.D. degree in Computer Science from the University of Washington. Liu's research interests are in computer graphics and robotics, including physics-based animation, character animation, optimal control, reinforcement learning, and computational biomechanics. She developed computational approaches to modeling realistic and natural human movements, learning complex control policies for humanoids and assistive robots, and advancing fundamental numerical simulation and optimal control algorithms. The algorithms and software developed in her lab have fostered interdisciplinary collaboration with researchers in robotics, computer graphics, mechanical engineering, biomechanics, neuroscience, and biology. Liu received a National Science Foundation CAREER Award, an Alfred P. Sloan Fellowship, and was named Young Innovators Under 35 by Technology Review. In 2012, Liu received the ACM SIGGRAPH Significant New Researcher Award for her contribution in the field of computer graphics.
\end{IEEEbiography}

\begin{IEEEbiography}[{\includegraphics[width=1in,height=1.25in,clip,keepaspectratio]{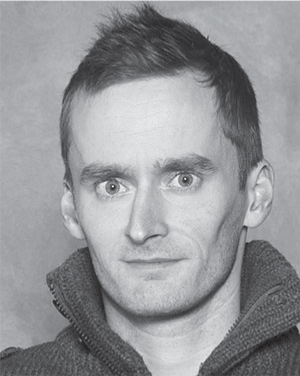}}]{Perttu H{\"a}m{\"a}l{\"a}inen}
  received an M.Sc.(Tech) degree from Helsinki University of Technology in 2001, an M.A. degree from the University of Art and Design Helsinki in 2002, and a doctoral degree in computer science from Helsinki University of Technology in 2007. Presently, Hämäläinen is an associate professor at Aalto University, publishing on human-computer interaction, computer animation, machine learning, and game research. Hämäläinen is passionate about human movement in its many forms, ranging from analysis and simulation to first-hand practice of movement arts such as parkour or contemporary dance.
\end{IEEEbiography}




\clearpage

\appendices

\section{Hyperparameters}\label{appendix:hyperparameters}
The hyperparameters used in the experiments are shown in Table \ref{table:parameters}. For the sake of better reproducability, almost all of the CMA-ES, PPO, and SAC parameters were initialized to their default values in \textit{cma} and \textit{stable-baselines} packages. Inspired by previous work in physically-based environments, we only increased  PPO's iteration experience budget from $128$ to $20000$ and $2000$ for humanoid and non-humanoid environments, respectively.

\begin{table}[b]
\centering
\caption{Hyperparameters used in the experiments.}
\begin{tabular}{|>{\centering\arraybackslash}p{2cm}|>{\centering\arraybackslash}p{4cm}|>{\centering\arraybackslash}p{1.0cm}|}
\hline
\textbf{Method} & \textbf{Parameter}	& \textbf{Value} \\ \hlineB{3.5}
\multirow{2}{2cm}{\centering Offline Optimization} & Number of iterations & $200$\\ \cline{2-3}
& Population size & $32$ \\ \hlineB{2.5}
\multirow{2}{2cm}{\centering Online Optimization} & Simulation horizon (seconds) & $2$\\ \cline{2-3}
& Population size & $250$ \\ \hlineB{2.5}
\multirow{12}{2cm}{\centering Reinforcement Learning (PPO)} & Discount factor $\left(\gamma\right)$ & $0.99$\\ \cline{2-3}
& Learning rate & $10^{-4}$\\ \cline{2-3}
& Number of mini batches & $4$\\ \cline{2-3}
& Number of optimization epochs & $4$\\ \cline{2-3}
& Value function clipping value & $0.2$\\ \cline{2-3}
& Iteration experience budget (humanoid) & $20000$ \\ \cline{2-3}
& Iteration experience budget (halfcheetah, hopper, and walker2d) & $2000$ \\ \cline{2-3}
& Entropy coefficient for the loss & $0.01$ \\ \cline{2-3}
& Value function coefficient for the loss & $0.5$ \\ \cline{2-3}
& Gradient clipping value & $0.5$ \\ \hlineB{2.5}
\multirow{18}{2cm}{\centering Reinforcement Learning (SAC)} & Discount factor $\left(\gamma\right)$ & $0.99$\\ \cline{2-3}
& Learning rate & $3 \times 10^{-4}$\\ \cline{2-3}
& Size of the replay buffer & $50000$\\ \cline{2-3}
& Number of transitions before learning starts & $100$\\ \cline{2-3}
& Training frequency per step & $1$\\ \cline{2-3}
& Target network update frequency per step & $1$\\ \cline{2-3}
& Batch size & $64$\\ \cline{2-3}
& Soft update coefficient $\tau$ & $0.005$\\ \cline{2-3}
& Number of gradient steps after each step & $1$\\ \cline{2-3}
& Random exploration probability & $0$\\ \hline
\end{tabular}
\vfill
\label{table:parameters}
\end{table}

\section{Single target vs. multi target}\label{appendix:convergence_curves}

In the experiments of the paper body, all the LLCs are queried with a sequence of $H$ target states. For completeness, we also tested a version where a single target state was given to be reached $H$ steps in the future. Figure \ref{fig:comparing_hs} shows the aggregated normalized scores of such single-target LLCs, augmenting Fig. 3 of the paper body.

\section{Convergence Curves}\label{appendix:convergence_curves}
To augment the discussion in Section 7.3, Figures \ref{fig:convergence_to}-\ref{fig:convergence_td3} show how an optimization iteration's mean episode/trajectory returns evolve with respect to the total simulation steps taken, with each agent, task, and action space. The graphs show the mean and standard deviations over ten training runs with different random seeds.

\section{Optimization Landscapes}\label{appendix:convergence_curves}
In Section 7.4, we visualized how using the LLC leads to smoother landscapes with wider high-return basins. In policy optimization, we have a slightly different explanation. Let us consider the case of a locomotion policy that keeps the character on a periodic movement trajectory. Now, consider perturbing some on-policy state, and how the actions that keep the character on the policy trajectory change with respect to the state perturbation. If one assumes an ideal LLC, the actions required to keep the character on the policy trajectory require no changes, whereas without an LLC, increasingly larger corrective actions are needed with larger state perturbations. Thus, the mapping from states to actions that the policy needs to learn is more complex without the LLC, and perturbing the policy network parameters from the optimum in different directions should have larger and less consistent effects on the policy return. 

To test the hypothesis above, we visualized the optimization landscape for reinforcement learning using PPO. In this setting, the 2D basis vectors were randomized in the policy neural network's weight space. For each point in the 2D grid, we ran the corresponding policy for $10$ episodes and averaged the episode returns. An example of the optimization landscapes is shown in Fig. \ref{fig:landscape_plots_ppo}. The landscapes using the low-level controller have slightly larger high-reward areas and also less false optima in the case of the half cheetah slow walk. However, the differences between the actions spaces are less pronounced than in trajectory optimization.  

\section{Ablation study: Feasible and Infeasible Target Trajectories}

As explained in Section 5.3, Algorithm 2 trains the LLCs using a mixture of feasible and infeasible states as the target trajectories in each episode. The infeasible target trajectories are added in the LLC training to ensure robustness to imperfect HLCs. For example, if an HLC is trained using RL, it will typically generate highly noisy and infeasible target trajectories in the initial training, before converging. To guarantee as smooth HLC optimization landscape as possible, the LLC should try to drive the agent towards even clearly infeasible target states, rather than producing random movements. The latter can easily happen if the LLC is only trained with feasible trajectories, as the output of a deep neural network can behave erratically when its input moves outside the training distribution.

To test the importance of the infeasible target trajectories, we conducted an ablation study with the infeasible trajectories removed from the LLC training. Figure \ref{fig:ablation_study} shows the results of this study using three environments (\textit{HalfCheetah-v2}, \textit{Walked2d-v2}, and \textit{Hopper-v2}) and the tasks explained in Section 4. Overall, the results are in line with the non-ablated ones, with no major difference in performance. Although this suggests that the infeasible target trajectories are not really needed with our HLC test tasks, we believe it is safer to include them in the general case, as they impose no extra computing cost and do not negatively impact the results.


\begin{figure*}[!htb]
\centering
{{\includegraphics[width=1.0\linewidth]{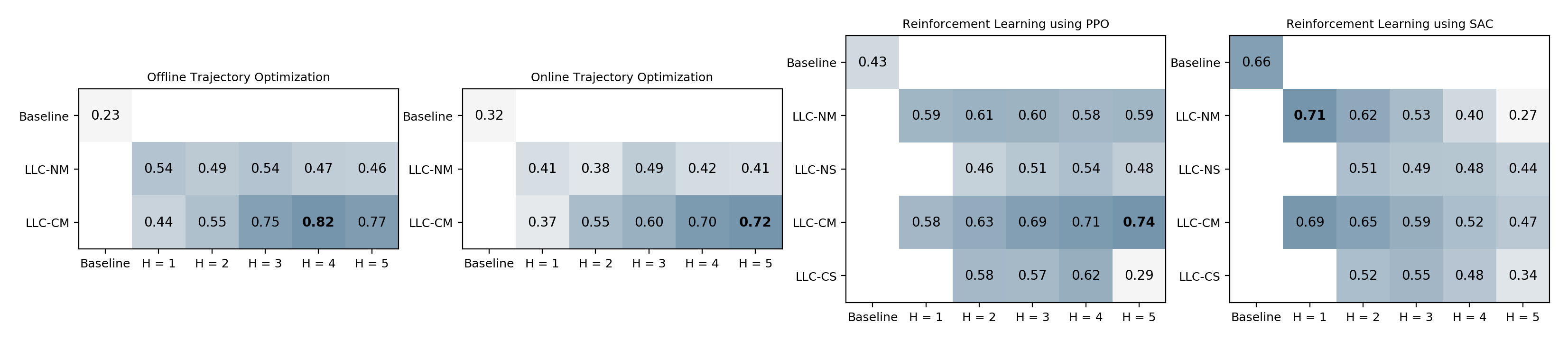} }}
\caption{Comparing different action spaces (baseline vs. LLC versions) and $H$ values in terms of normalized average episode/trajectory returns (higher is better) in the four optimization approaches. The "N" and "C" suffixes correspond to LLCs trained with naive and contact-based exploration, respectively. The "S" and "M" distinguish between LLCs with single and multiple target states. }
\label{fig:comparing_hs}
\end{figure*}

\begin{figure*}[!ht]
\centering
{\includegraphics[width=0.9\linewidth]{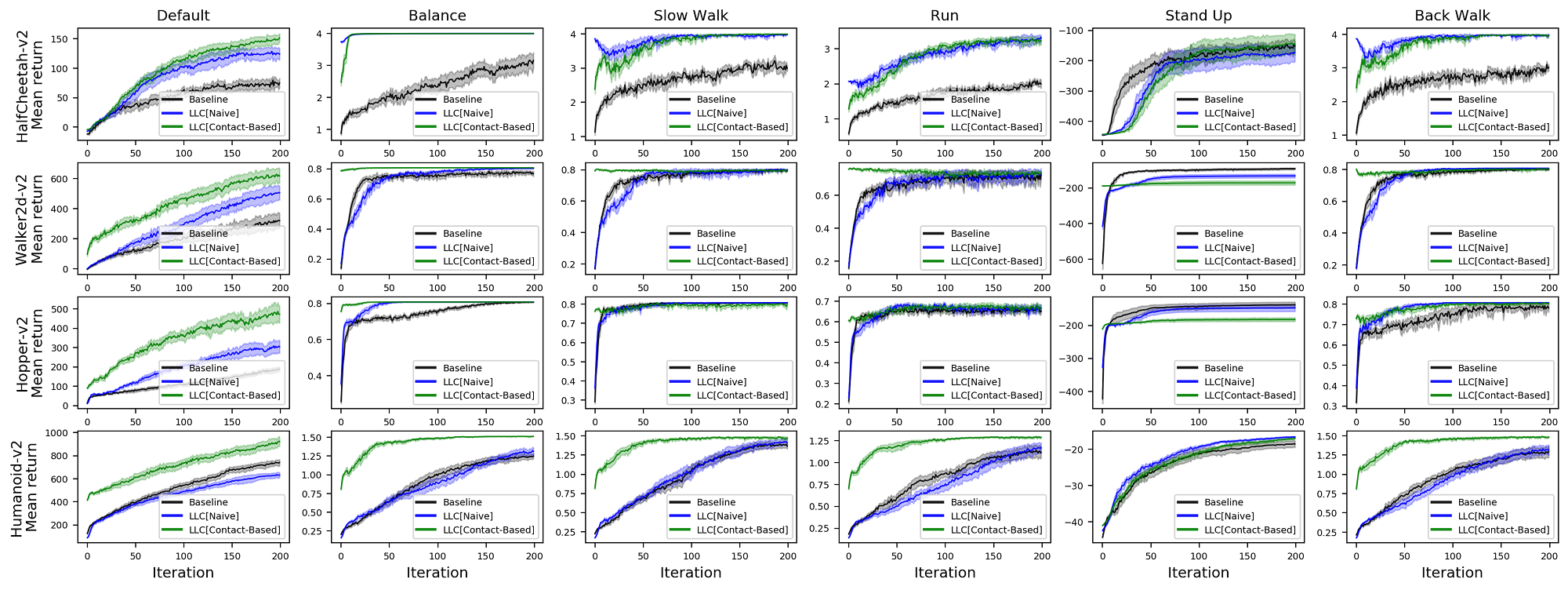} }
\caption{Convergence plot of mean trajectory return in offline trajectory optimization with LLC horizon $H=4$.}
\label{fig:convergence_to}
\end{figure*}

\begin{figure*}[!ht]
\centering
{\includegraphics[width=0.9\linewidth]{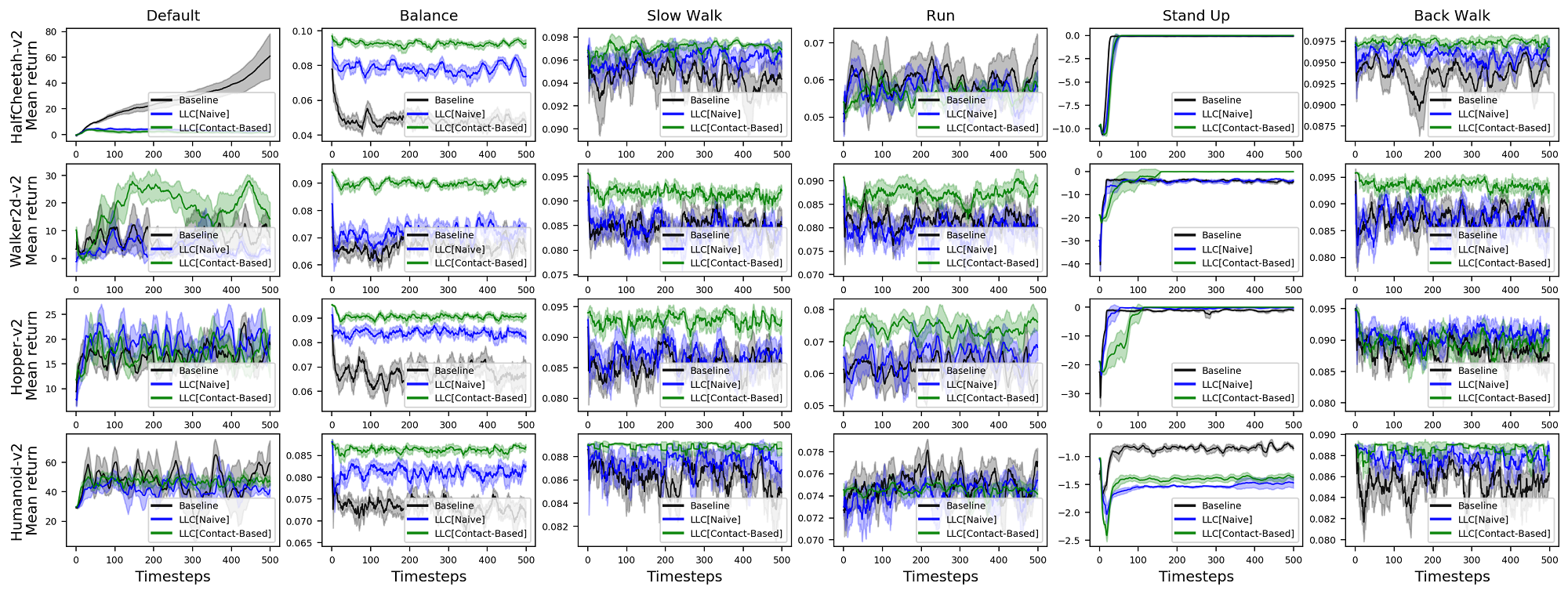}}
\caption{Convergence plot of mean trajectory return in online trajectory optimization with LLC horizon $H=5$.}
\label{fig:convergence_oo}
\end{figure*}

\begin{figure*}[!ht]
\centering
{\includegraphics[width=0.9\linewidth]{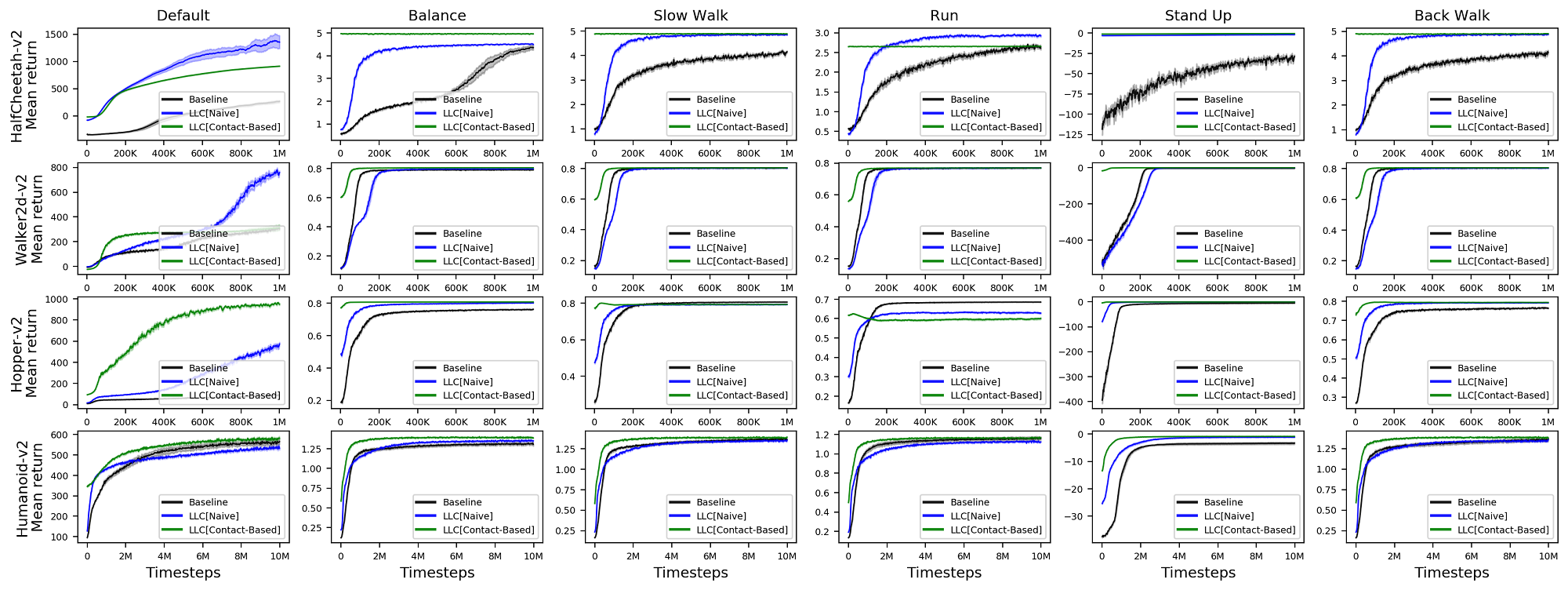} }
\caption{Convergence plot of mean episode return in reinforcement learning using PPO with LLC horizon $H=5$.}
\label{fig:convergence_ppo}
\end{figure*}

\begin{figure*}[!ht]
\centering
{\includegraphics[width=0.9\linewidth]{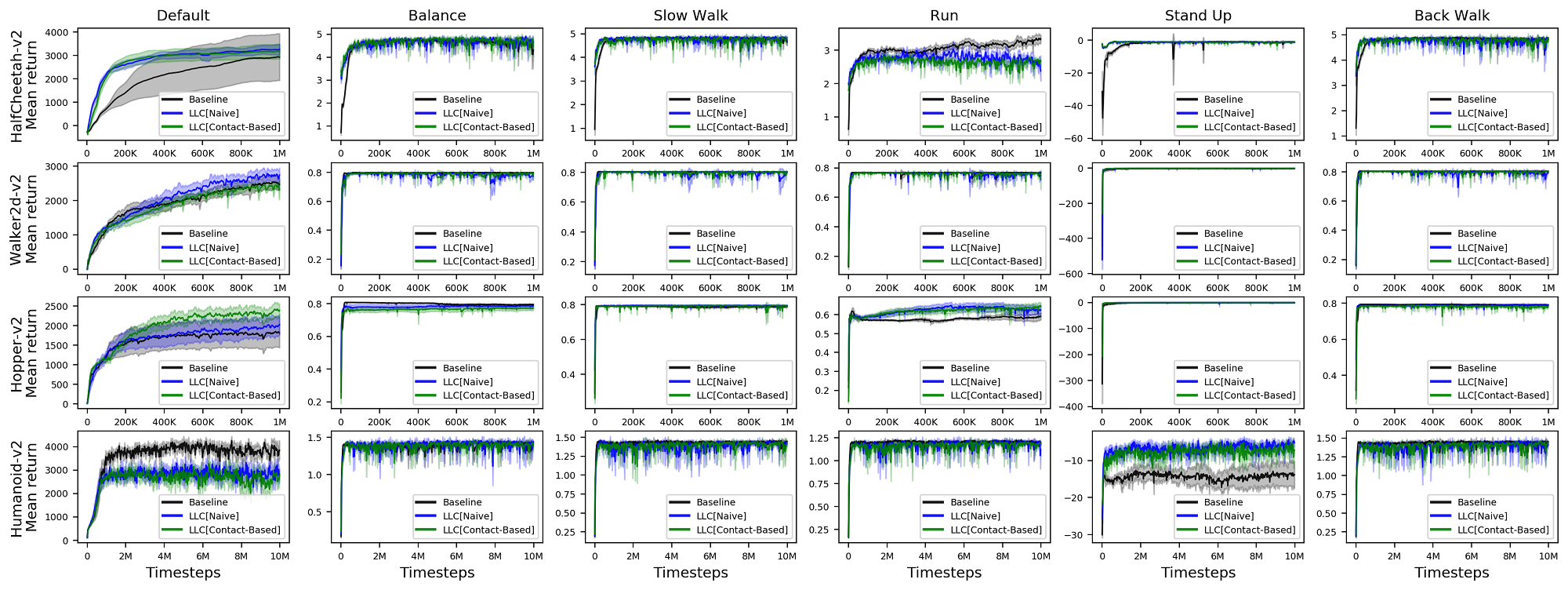} }
\caption{Convergence plot of mean episode return in reinforcement learning using SAC with LLC horizon $H=1$.}
\label{fig:convergence_sac}
\end{figure*}

\begin{figure*}[!ht]
\centering
{\includegraphics[width=0.9\linewidth]{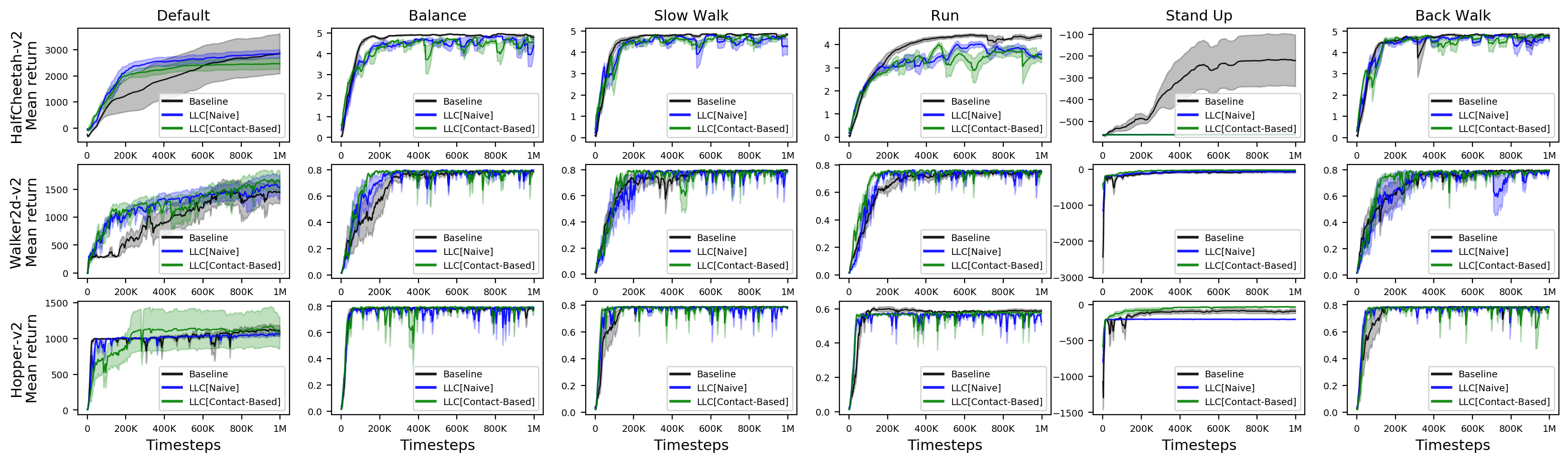} }
\caption{Convergence plot of mean episode return in reinforcement learning using TD3 with LLC horizon $H=1$.}
\label{fig:convergence_td3}
\end{figure*}

\begin{figure*}[!t]
\centering
{\includegraphics[width=0.775\linewidth]{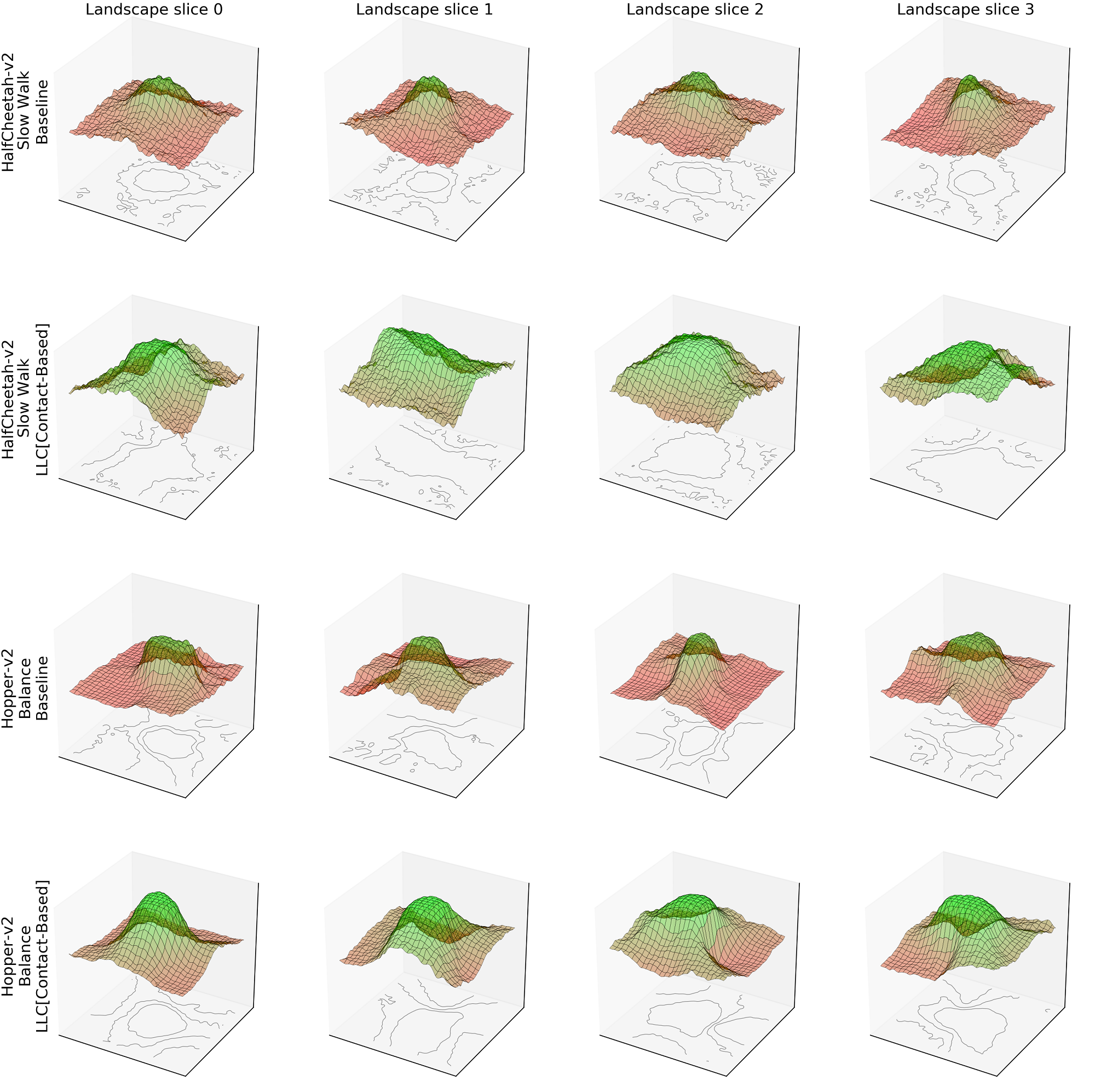}}
\caption{PPO optimization landscapes of HalfCheetah-v2 slow walking and Hopper-v2 balancing with/without the low-level controllers (the up-axis shows the average episode return, i.e., higher is better). Using low-level controllers leads to slightly smoother landscapes with higher returns.}
\label{fig:landscape_plots_ppo}
\end{figure*}

\begin{figure*}[!ht]
\centering
{{\includegraphics[width=1.0\linewidth]{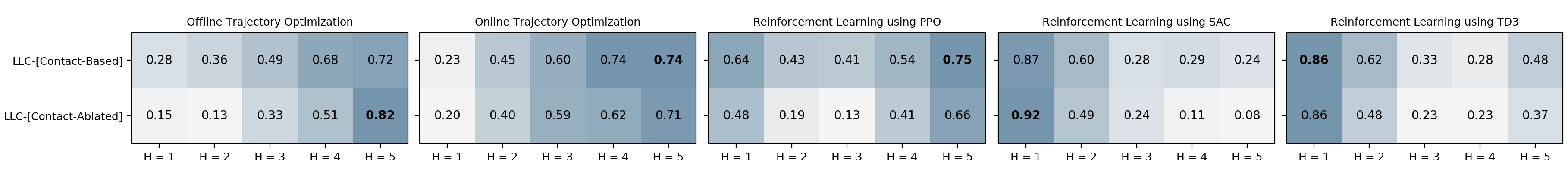} }}
\caption{Results of the ablation study, where \textit{LLC[Contact-Ablated]} results are obtained using LLCs trained only with feasible target state trajectories.} 
\label{fig:ablation_study}
\end{figure*}

\end{document}